\documentclass{article}

    \usepackage[final]{neurips_2020}

\usepackage[utf8]{inputenc} 
\usepackage[T1]{fontenc}    
\usepackage{url}            
\usepackage{booktabs}       
\usepackage{amsfonts}       
\usepackage{nicefrac}       
\usepackage{microtype}      

\usepackage{tikz,pgfplots}
\usepackage{natbib}
\usepackage{graphicx,amsthm,wrapfig}
\usepackage{multirow,tikz}
\usepackage{mathrsfs}
\usepackage{mdsymbol,comment}
\usepackage{algorithm}
\usepackage{algorithmic}
\usepackage{threeparttable}
\usepackage{footnote}
\usepackage{multicol}
\usepackage{comment}
\usepackage{enumitem}
\usepackage{amsmath}
\usepackage{siunitx}
\usepackage{makecell}
\usepackage{subcaption}

\usepackage{fdsymbol}

\makeatletter
\newtheorem*{rep@theorem}{\rep@title}
\newcommand{\newreptheorem}[2]{%
\newenvironment{rep#1}[1]{%
 \def\rep@title{#2 \ref{##1}}%
 \begin{rep@theorem}}%
 {\end{rep@theorem}}}
\makeatother

\newreptheorem{lemma}{Lemma}
\newreptheorem{theorem}{Theorem}
\newreptheorem{proposition}{Proposition}

\usetikzlibrary{fit,matrix,positioning, 
decorations.pathreplacing,calc,decorations,arrows,shadows,patterns}

\usetikzlibrary{arrows,fit,positioning,shapes,matrix}

\DeclareMathAlphabet{\mathcal}{OMS}{cmsy}{m}{n}
\SetMathAlphabet{\mathcal}{bold}{OMS}{cmsy}{b}{n}

\newenvironment{nospaceflalign*}
 {\setlength{\abovedisplayskip}{1pt}\setlength{\belowdisplayskip}{1pt}%
  \csname flalign*\endcsname}
 {\csname endflalign*\endcsname\ignorespacesafterend}
 
 \newenvironment{nospaceflalign}
 {\setlength{\abovedisplayskip}{1pt}\setlength{\belowdisplayskip}{1pt}%
  \csname flalign\endcsname}
 {\csname endflalign\endcsname\ignorespacesafterend}

\newcommand{\Feng}[1]{{\color{black} #1}}
\newtheorem{example-set}{Example}
\newtheorem{theorem}{Theorem}
\newtheorem{definition}{Definition}

\newtheorem{lemma}{Lemma}
\newtheorem{proposition}{Proposition}
\newtheorem{problem definition}{Problem Definition}

\newdimen\arrowsize
\pgfarrowsdeclare{arcsq}{arcsq}
{
	\arrowsize=0.2pt
	\advance\arrowsize by .5\pgflinewidth
	\pgfarrowsleftextend{-4\arrowsize-.5\pgflinewidth}
	\pgfarrowsrightextend{.5\pgflinewidth}
}
{
	\arrowsize=0.8pt
	\advance\arrowsize by .5\pgflinewidth
	\pgfsetdash{}{0pt} 
	\pgfsetroundjoin   
	\pgfsetroundcap    
	\pgfpathmoveto{\pgfpoint{0\arrowsize}{0\arrowsize}}
	\pgfpatharc{-90}{-140}{4\arrowsize}
	\pgfusepathqstroke
	\pgfpathmoveto{\pgfpointorigin}
	\pgfpatharc{90}{140}{4\arrowsize}
	\pgfusepathqstroke
}

\title{Generalized Independent Noise Condition
\\ for Estimating Latent Variable Causal Graphs}

\author{ \hspace{-3.5mm}
\textbf{Feng Xie}\thanks{These authors contributed equally to this work. The work was done while FX was visiting CMU.} \hspace{0.02mm} $^{1,2}$, \textbf{Ruichu Cai}$^{*}$$^{1,3}$, \textbf{Biwei Huang}$^{4}$, \textbf{Clark Glymour}$^4$, \textbf{Zhifeng Hao}$^{1,5}$, \textbf{Kun Zhang}$^{*}$$^4$~~\\
$^1$ School of Computer Science, Guangdong University of Technology, Guangzhou, China\\
$^2$ School of Mathematical Sciences, Peking University, Beijing, China\\
$^3$ Pazhou Lab, Guangzhou, China\\
$^4$ Department of Philosophy, Carnegie Mellon University, Pittsburgh, USA\\
$^5$ 
School of Mathematics and Big Data, Foshan University, Foshan, China\\
\texttt{xiefeng009@gmail.com, cairuichu@gdut.edu.cn, biweih@andrew.cmu.edu}\\
\texttt{cg09@andrew.cmu.edu, zfhao@gdut.edu.cn, kunz1@cmu.edu}
}

\begin{document}

\maketitle
\begin{abstract}
Causal discovery aims to recover causal structures or models underlying the observed data. Despite its success in certain domains, most existing methods focus on causal relations between observed variables, while in many scenarios the observed ones may not be the underlying causal variables (e.g., image pixels), but are generated by latent causal variables or confounders that are causally related. To this end, in this paper, we consider Linear, Non-Gaussian Latent variable Models (LiNGLaMs), in which latent confounders are also causally related, and propose a Generalized Independent Noise (GIN) condition to estimate such latent variable graphs. Specifically, for two observed random vectors $\mathbf{Y}$ and $\mathbf{Z}$, GIN holds if and only if $\omega^{\intercal}\mathbf{Y}$ and $\mathbf{Z}$ are statistically independent, where $\omega$ is a parameter vector  characterized from the cross-covariance between 
$\mathbf{Y}$ and $\mathbf{Z}$. From the graphical view, roughly speaking, GIN implies that causally earlier latent common causes of variables in $\mathbf{Y}$ d-separate $\mathbf{Y}$ from $\mathbf{Z}$. Interestingly, we find that the independent noise condition, i.e., if there is no confounder, causes are independent from the error of regressing the effect on the causes, can be seen as a special case of GIN. Moreover, we show that GIN helps locate latent variables and identify their causal structure, including causal directions. We further develop a recursive learning algorithm to achieve these goals. Experimental results on synthetic and real-world data demonstrate the effectiveness of our method.
\end{abstract}
\vspace{-1mm}
\section{Introduction} 
\vspace{-1mm}
Identifying causal relationships from observational data, known as causal discovery, has drawn much attention in the fields of empirical science and artificial intelligence \citep{spirtesautomated, Pearl2019SevenTools}. Most causal discovery approaches focus on the situation without latent variables, such as the PC algorithm~\citep{spirtes1991PC}, Greedy Equivalence Search (GES)~\citep{chickering2002optimal}, and
methods based on the Linear, Non-Gaussian Acyclic Model (LiNGAM)~\citep{shimizu2006linear}, the Additive Noise Model (ANM)~\citep{hoyer2009ANM}, and the Post-NonLinear causal model (PNL)~\citep{Zhang06_iconip, zhang2009PNL}. 
However, although these methods have been used in a range of fields, they may fail to produce convincing results in cases with latent variables (or more specifically, confounders), because they do not properly take into account 
the influences from latent variables as well as many other practical issues \citep{NSR_review18}.

Causal discovery with latent variables has attracted much attention. Some approaches attempt to handle the question based on conditional independence constraints, including the FCI algorithm~\citep{spirtes1995causal}, RFCI~\citep{colombo2012learning}, and their variants. They focus on estimating the causal relationships between observed variables rather than that between latent variables. However, in real-world scenarios, it may not be the case---there are also causal relationships between latent variables. 
Later, it was shown that by utilizing vanishing Tetrad conditions \citep{spearman1928pearson} and, more generally, t-separation, one is able to identify latent variables in linear-Gaussian models \citep{Silva-linearlvModel, Sullivant-T-separation}. Furthermore, by leveraging an extended t-separation \citep{spirtes2013calculation-t-separation}, a more reliable and faster algorithm, called  FindOneFactorClusters (FOFC), was developed \citep{Kummerfeld2016}. However, these methods may not be able to identify causal directions between latent variables, and they require strong constraints that each latent variable should have at least three pure measurement variables.\Feng{\footnote{The variable is neither the cause nor the effect of other measurement variables.}} Such limitation is because they only rely on rank constraints on the covariance matrix, but fail to take into account higher-order statistics. 
To make use of higher-order information, one may apply overcomplete independent component analysis \citep{hoyer2008estimation, shimizu2009estimation}, but it does not consider the causal structure between latent variables and the size of the equivalence class of the identified structure could be very large \citep{entner2010discovering, tashiro2014parcelingam}. Another interesting work by \citet{anandkumar2013learning} extracts second-order statistics in identifying latent factors, while using non-Gaussianity when estimating causal relations between latent variables. \citet{Zhang17_IJCAI} and \citet{CDNOD_jmlr} considered a special type of confounders due to distribution shifts. 

Recently, a condition about a particular type of independence relationship between any three variables, called Triad condition, was proposed \citep{cai2019triad}, together with the LSTC algorithm to discover the structure between latent variables. 
Nevertheless, this method does not apply to the case where there are multiple latent variables behind two observed variables.

It is well known that one may use the independent noise condition to recover the causal structure from linear non-Gaussian data without latent variables~\citep{shimizu2011directlingam}. Then a question naturally rises: is it possible to solve the latent-variable problem, by introducing non-Gaussianity and a condition similar to the independent noise condition? 
Interestingly, we find that it can be achieved by testing the independence between $\omega^{\intercal} \mathbf{Y}$ and $\mathbf{Z}$, where $\mathbf{Y}$ and $\mathbf{Z}$ are two observed random vectors, and $\omega$ is a parameter vector based on the cross-covariance between $\mathbf{Y}$ and $\mathbf{Z}$. 
If $\omega^{\intercal} \mathbf{Y}$ and $\mathbf{Z}$ are statistically independent, we term this condition Generalized Independent Noise (GIN) condition. We show that the well-known independent noise condition can be seen as a special case of GIN. 
From the view of graphical models, roughly speaking, if the GIN condition holds, then in the Linear Non-Gaussian Latent variable Model (LiNGLaM), the causally earlier latent common causes of variables in $\mathbf{Y}$ d-separate $\mathbf{Y}$ from $\mathbf{Z}$. 
By leveraging GIN, we further develop a practical algorithm to identify important information of the LiNGLaM, including where the latent variables are, the number of latent variables behind any two observed variables, and the causal order of the latent variables. 

The contributions of this work are three-fold. 1) We define the GIN condition for an ordered pair of variables sets, provide mathematical conditions that are sufficient for it, and show that the independent noise condition can be seen as its special case. 2) We then further
establish a connection between the GIN condition and the graphical patterns in the LiNGLaM, including specific d-separation relations. 
3) We exploit GIN to estimate the LiNGLaM, which allows causal relationships between latent variables and multiple latent variables behind any two observed variables. Compared to existing work, a uniquely appealing feature of the proposed method is that it is able to identify the causal order of the latent variables and determine the number of latent variables behind any two observed variables. 

\vspace{-3mm}
\section{Problem Definition}
\vspace{-3mm}
In this paper, we focus on a particular type of linear acyclic latent variable causal models. We use $\mathbf{V}={\bf{X}} \cup {\bf{L}}$ to denote the total set of variables, where ${\bf{X}}$ denote the set of observed variables, with ${\bf{X}}= \{X_1,X_2, ...X_m\}$, and ${\bf{L}}$ denote the set of latent variables, with ${\bf{L}}= \{L_1,L_2, ...L_n\}$. 
We assume that any variable in $\mathbf{V}$ satisfy the following generating process:
${V_i} = \sum_{k(j)<k(i)}{{b_{ij}}{V_j} + {\varepsilon_{V_i}}},i = 1,2,...,m+n,$
where $k(i)$ represents the causal order of variables in a directed acyclic graph, so that no later variable causes any earlier variable, $b_{ij}$ denotes the causal strength from $V_j$ to $V_i$, and ${\varepsilon_{V_i}}$ are independent and identically distributed noise variables. Without loss of generality, we assume that all variables have a zero mean (otherwise can be centered). 
The definition of our model is given below.

\begin{definition}[Linear Non-Gaussian Latent Variable Model (LiNGLaM)]\label{def-model}
A LiNGLaM, besides linear and acyclic assumptions, has the following assumptions:
\begin{itemize}[noitemsep,topsep=-3pt,leftmargin=30pt]
\item[A1.] [Measurement Assumption] There is no observed variable in $\mathbf{X}$ being an ancestor of any latent variables in $\mathbf{L}$.{\footnote{\Feng{Here, this assumption follows the definition in \citet{Silva-linearlvModel} and it is equivalent to say that there is no observed variable in $\mathbf{X}$ being an parent of any latent variables in $\mathbf{L}$}.}} 
\item[A2.] [Non-Gaussianity Assumption] The noise terms are non-Gaussian.
\item[A3.] [Double-Pure Child Variable Assumption] Each latent variable set $\mathbf{L'}$, in which every latent variable directly causes the same set of observed variables, has at least $2\textrm{Dim}(\mathbf{L'})$ pure measurement variables as children.{\footnote{\Feng{$2\textrm{Dim}(\mathbf{L'})$ denotes 2 times the dimension of $\mathbf{L'}$}.}}
\item[A4.] [Purity Assumption] There is no direct edge between observed variables.
\end{itemize}
\end{definition}
\vspace{-1mm}
\begin{wrapfigure}{r}{0.44\textwidth} \vspace{-.8cm}
   \setlength{\abovecaptionskip}{0pt}
	\setlength{\belowcaptionskip}{-6pt}
	\begin{center}
		\begin{tikzpicture}[scale=1.2, line width=0.5pt, inner sep=0.2mm, shorten >=.1pt, shorten <=.1pt]
		\draw (1.5, 2.5) node(i-L1) [circle, draw] {{\footnotesize\,$L_1$\,}};
		\draw (1.5, 0.5) node(i-L2) [circle, draw] {{\footnotesize\,$L_2$\,}};
		\draw (3.5, 2.3) node(i-L3) [circle, draw] {{\footnotesize\,$L_3$\,}};
		\draw (3.5,0.7) node(i-L4) [circle, draw] {{\footnotesize\,$L_4$\,}};
		\draw (0.5, 1.5) node(i-X1) [] {{\footnotesize\,$X_1$\,}};
		\draw (1, 1.5) node(i-X2) [] {{\footnotesize\,$X_2$\,}};
		\draw (2, 1.5) node(i-X3) [] {{\footnotesize\,{$X_3$}\,}};
		\draw (2.5, 1.5) node(i-X4) [] {{\footnotesize\,{$X_4$}\,}};
		\draw (4.5, 2.8) node(i-Y1) [] {{\footnotesize\,{$X_5$}\,}};
		\draw (4.5, 1.8) node(i-Y3) [] {{\footnotesize\,{$X_6$}\,}};
		\draw (4.5, 1.2) node(i-Z1) [] {{\footnotesize\,{$X_7$}\,}};
		\draw (4.5, 0.2) node(i-Z3) [] {{\footnotesize\,{$X_{8}$}\,}};
		%
		\draw[-arcsq] (i-L1) -- (i-X1) node[pos=0.5,sloped,above] {\scriptsize{$a_{1}$}}; 
		\draw[-arcsq] (i-L1) -- (i-X2)node[pos=0.5,sloped,below] {\scriptsize{$a_{2}$}}; 
		\draw[-arcsq] (i-L1) -- (i-X3) node[pos=0.5,sloped,below] {\scriptsize{$a_{3}$}};
		\draw[-arcsq] (i-L1) -- (i-X4) node[pos=0.5,sloped,above] {\scriptsize{$a_{4}$}};
		\draw[-arcsq] (i-L2) -- (i-X1) node[pos=0.5,sloped,below] {\scriptsize{$b_{1}$}}; 
		\draw[-arcsq] (i-L2) -- (i-X2)node[pos=0.5,sloped,above] {\scriptsize{$b_{2}$}}; 
		\draw[-arcsq] (i-L2) -- (i-X3) node[pos=0.5,sloped,above] {\scriptsize{$b_{3}$}};
		\draw[-arcsq] (i-L2) -- (i-X4) node[pos=0.5,sloped,below] {\scriptsize{$b_{4}$}};
		\draw[-arcsq] (i-L3) -- (i-Y1) node[pos=0.5,sloped,above] {\scriptsize{$c_{1}$}}; 
		\draw[-arcsq] (i-L3) -- (i-Y3)node[pos=0.5,sloped,below] {\scriptsize{$c_{2}$}}; 
		\draw[-arcsq] (i-L4) -- (i-Z1) node[pos=0.5,sloped,above] {\scriptsize{$d_{1}$}}; 
		\draw[-arcsq] (i-L4) -- (i-Z3)node[pos=0.5,sloped,below] {\scriptsize{$d_{2}$}}; 
		\draw[-arcsq] (i-L1) -- (i-L2) node[pos=0.5,sloped,above] {\scriptsize{$\alpha$}};
		\draw [-arcsq] (i-L1) edge[bend right=-25] (i-L4);
		\draw (2.5,2.2) node(label-iii) [] {{\scriptsize\,$\gamma$\,}};
		\draw [-arcsq] (i-L1) edge[bend right=-25] (i-L3);
		\draw (2.5,2.8) node(label-iii) [] {{\scriptsize\,$\beta$\,}};
		\draw [-arcsq] (i-L2) edge[bend right=25] (i-L3);
		\draw (2.5,0.8) node(label-iii) [] {{\scriptsize\,$\sigma$\,}};
		\draw [-arcsq] (i-L2) edge[bend right=25] (i-L4);
		\draw (2.5,0.4) node(label-iii) [] {{\scriptsize\,$\eta$\,}};
		\draw[-arcsq] (i-L3) -- (i-L4) node[pos=0.5,sloped,above] {\scriptsize{$\theta$}};
		\end{tikzpicture}
		\caption{A causal structure involving $4$ latent variables and 8 observed variables, where each pair of observed variables in $\{X_1,X_2,X_3,X_4\}$ are affected by two latent variables.}
		\vspace{-0.4cm}
		\label{fig:simple-example-explain-GIN-constraint} 
	\end{center}
\end{wrapfigure}
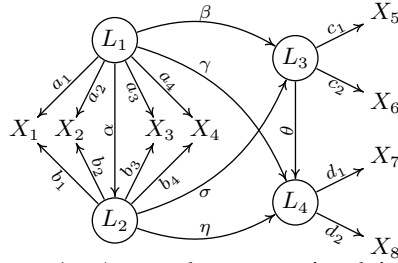
The key difference to existing researches considering linear latent models, such as ~\citet{bollen1989structural,Silva-linearlvModel}, is that we introduce the assumptions A2$\sim$A4, allowing us to identify the casual structure over latent variables, including casual directions.
\Feng{Figure 1 shows a simple example that satisfies the LiNGLaM}.
For \emph{Non-Gaussianity Assumption}, the non-Gaussian distribution are expected to be ubiquitous, due to Cram\'{e}r Decomposition Theorem \citep{Cramer62}, as stated in \citet{spirtes2016causal}.
Notice that the \emph{Double-Pure Child Variable Assumption} is much milder than that in Tetrad-based methods: for latent variable set $\mathbf{L'}$, we only need $2\textrm{Dim}(\mathbf{L'})$ pure observed variables, while Tetrad needs $2\textrm{Dim}(\mathbf{L'})+1$ pure observed variables. 
In Section 6, we will briefly discuss the situation where Assumption A4 is violated.

\vspace{-3mm}
\section{GIN Condition and Its Implications in LiNGLaM}
\vspace{-3mm}
In this section, we first briefly review the Independent Noise (IN) condition in linear non-Gaussian causal models with no latent variables. Then we formulate the Generalized Independent Noise (GIN) condition and show that it contains the independent noise condition as a special case. We further illustrate how GIN is applied to identify causal relations between latent variables of any two considered groups of observed variables. Finally, we present theoretical results regarding the graphical implications of the GIN condition, which can be used to discover latent variable structures. 
\vspace{-3mm}
\subsection{Independent Noise Condition in Functional Causal Models}
\vspace{-3mm}
Below, we give the independent noise condition, which has been used in causal discovery of linear, non-Gaussian networks without confounders (e.g., in \citet{shimizu2011directlingam}).
\begin{definition}[IN condition]
Let $Y$ be a single variable and $\mathbf{Z}$ a set of variables. Suppose all variables follow the linear non-Gaussian acyclic causal model and are observed.  We say that $(\mathbf{Z},Y)$ follows the IN condition, if and only if the residual of regressing $Y$ on $\mathbf{Z}$ is statistically independent from $\mathbf{Z}$. Mathematically, let $\tilde{\omega}$ be the vector of regression coefficients, that is, $\tilde{\omega}\coloneqq \mathbb{E}[Y \mathbf{Z}^\intercal]\mathbb{E}^{-1}[\mathbf{Z} \mathbf{Z}^\intercal]$; the IN condition holds for $(\mathbf{Z},Y)$ if and only if $\tilde{E}_{Y||\mathbf{Z}} = Y - \tilde{\omega}^{\intercal}\mathbf{Z}$ is independent from $\mathbf{Z}$.
\end{definition}
\vspace{-1mm}

Lemma 1 in \citet{shimizu2011directlingam} considers the case where $\mathbf{Z}$ is a single variable and shows that $(\mathbf{Z},Y)$ satisfies the IN condition if and only if $\mathbf{Z}$ is an exogenous (or root) variable relative to $Y$, based on which one can identify the causal relation between $Y$ and $\mathbf{Z}$. As a direct extension of this result, we show that in the case where $\mathbf{Z}$ contains multiple variables, $(\mathbf{Z},Y)$ satisfies the IN condition if and only if all variables in $\mathbf{Z}$ are causally earlier than $Y$ and there is no common cause behind any variable in $\mathbf{Z}$ and $Y$. \Feng{This result is given in the following Proposition.
\begin{proposition}\label{pro-extension-IN}
Suppose all considered variables follow the linear non-Gaussian acyclic causal model and are observed. Let $\mathbf{Z}$ be a subset of those variables and $Y$ be a single variable. Then the following statements are equivalent.
\begin{itemize}[noitemsep,topsep=-1pt]
    \item[(A)] 1) All variables in $\mathbf{Z}$ are causally earlier than $Y$,  and 2) there is no common cause for each variable in $\mathbf{Z}$ and $Y$ that is not in $\mathbf{Z}$.
    \item[(B)] $(\mathbf{Z},Y)$ satisfies the IN condition.
\end{itemize}
\end{proposition}
}
\vspace{-1mm}
\subsection{Generalized Independent Noise Condition}
\vspace{-1mm}
Below, we first give the definition of the GIN condition, followed by an illustrative example.
\begin{definition}[GIN condition]
Let $\mathbf{Y}$ and $\mathbf{Z}$ be two observed random vectors. Suppose the variables follow the linear non-Gaussian acyclic causal model. Define the surrogate-variable of $\mathbf{Y}$ relative to $\mathbf{Z}$, as
\begin{nospaceflalign}\label{eq-E_XZ}
E_{\mathbf{Y}||\mathbf{Z}} \coloneqq \omega^\intercal \mathbf{Y},
\end{nospaceflalign}
where $\omega$ satisfies $\omega^\intercal \mathbb{E}[\mathbf{Y}\mathbf{Z}^\intercal] = \mathbf{0}$ and $\omega \neq \mathbf{0}$.  
We say that $(\mathbf{Z},\mathbf{Y})$ follows the GIN condition if and only if $E_{\mathbf{Y}||\mathbf{Z}}$ is independent from $\mathbf{Z}$.
\end{definition}
In other words, $(\mathbf{Z},\mathbf{Y})$ violates the GIN condition if and only if $E_{\mathbf{Y}||\mathbf{Z}}$ is dependent on $\mathbf{Z}$. \Feng{Notice that the Triad condition~\citep{cai2019triad} can be seen as a restrictive, special case of the GIN condition, where $\textrm{Dim}(\mathbf{Y})=2$ and $\textrm{Dim}(\mathbf{Z})=1$.}
We give an example to illustrate that there is a connection between this condition and the causal structure. According to the structure in Figure \ref{fig:simple-example-explain-GIN-constraint} and by assuming faithfulness, we have that $(\{X_4,X_5\},\{X_1,X_2,X_3\})$ satisfies the GIN condition, as explained below. The causal models of latent variables is $L_1 =\varepsilon_{L_1}$, $L_2 =\alpha L_1+\varepsilon_{L_2}=\alpha \varepsilon_{L_1}+ \varepsilon_{L_2}$, and $L_3 =\beta L_1+\sigma L_2+ \varepsilon_{L_3}=(\beta+\alpha\sigma) \varepsilon_{L_1}+ \sigma\varepsilon_{L_2}+\varepsilon_{L_3}$, and 
$\{X_1,X_2,X_3\}$ and $\{X_4,X_5\}$ can then be represented as
\begin{nospaceflalign}
\underbrace{\left[\begin{matrix}
{{X_1}}\\
{{X_2}}\\
{{X_3}}
\end{matrix}\right]}_{\mathbf{Y}} & =  {\left[\begin{matrix}
a_1 & {b_1}\\
a_2 & {b_2}\\
a_3 & {b_3}
\end{matrix}\right]}\left[\begin{matrix}
{{{{L_1}}}}\\
{{{{L_2}}}}
\end{matrix}\right] + \underbrace{\left[\begin{matrix}
{{\varepsilon_{{X_1}}}}\\
{{\varepsilon_{{X_2}}}}\\
{{\varepsilon_{{X_3}}}}
\end{matrix}\right]}_{\mathbf{E_Y}},~~~~~~~
\underbrace{\left[\begin{matrix}
{{X_4}}\\
{{X_5}}
\end{matrix}\right]}_{\mathbf{Z}} =  {\left[\begin{matrix}
a_4 &  b_4\\
\beta c_1 &  \sigma c_1
\end{matrix}\right]}\left[\begin{matrix}
{{L_{1}}}\\
{{L_{2}}}
\end{matrix}\right] + \underbrace{\left[\begin{matrix}
{{\varepsilon_{{X_4}}}}\\
{{\varepsilon_{{X^{'}_5}}}}
\end{matrix}\right]}_{\mathbf{E_Z}}, \label{equ-y1-y2}
\end{nospaceflalign}
where $\varepsilon_{X^{'}_5}=c_{2}\varepsilon_{L_3}+\varepsilon_{X_5}$.
According to the above equations,  $\omega^\intercal \mathbb{E}[\mathbf{Y}\mathbf{Z}^\intercal] = \mathbf{0} \Rightarrow \omega = [a_2b_3-b_2a_3,b_1a_3-a_1b_3, a_1b_2-b_1a_2]^{\intercal}$. Then we can see  $E_{\mathbf{Y}||\mathbf{Z}} = \omega^{\intercal}\mathbf{Y}= \omega^{\intercal}\mathbf{E_Y}$, and further because $\mathbf{E_Y} \upvDash \mathbf{Z}$, we have $E_{\mathbf{Y}||\mathbf{Z}} \upvDash \mathbf{Z}$. That is to say, $(\{X_4,X_5\},\{X_1,X_2,X_3\})$ satisfies the GIN condition. Intuitively, we have  $E_{\mathbf{Y}||\mathbf{Z}} \upvDash \mathbf{Z}$ because although $\{X_1,X_2,X_3\}$ were generated by $\{L_1,L_2\}$, which are not measurable, $E_{\mathbf{Y}||\mathbf{Z}}$, as a particular linear combination of $\mathbf{Y} = \{X_1,X_2,X_3\}$, successfully removes the influences of $\{L_1,L_2\}$ by properly making use of $\mathbf{Z} = \{X_4,X_5\}$ as a ``surrogate". 


Next, we discuss a situation where GIN is violated. For example, in this structure, $(\{X_3,X_6\},\{X_1,X_2,X_5\})$ violates GIN. Specifically, the corresponding variables satisfy the following equations:
\begin{nospaceflalign}
\underbrace{\left[\begin{matrix}
{{X_1}}\\
{{X_2}}\\
{{X_5}}
\end{matrix}\right]}_{\mathbf{Y}} & = \left[\begin{matrix}
{a_1} & {b_1} \\
{a_2} & {b_2} \\
\beta c_1 & \sigma c_2 
\end{matrix}\right]\left[\begin{matrix}
{{L_{1}}}\\
{{L_{2}}}
\end{matrix}\right] + \underbrace{\left[\begin{matrix}
{{\varepsilon _{{X_1}}}}\\
{{\varepsilon _{{X_2}}}}\\
{{\varepsilon _{{X^{'}_5}}}}
\end{matrix}\right]}_{\mathbf{E_Y}},~~~~~~~
\underbrace{\left[\begin{matrix}
{{X_3}}\\
{{X_6}}
\end{matrix}\right]}_{\mathbf{Z}} = \left[\begin{matrix}
{a_3} & {b_3}\\
\beta c_2 &  \sigma c_2
\end{matrix}\right]\left[\begin{matrix}
{{L_{1}}}\\
{{L_{2}}}
\end{matrix}\right] + \underbrace{\left[\begin{matrix}
{{\varepsilon _{{X_3}}}}\\
{\varepsilon _{X^{'}_6}}
\end{matrix}\right]}_{\mathbf{E_Z}},
\end{nospaceflalign}
where $\varepsilon_{X^{'}_6}=c_{2}\varepsilon_{L_3}+\varepsilon_{X_6}$.
Then under faithfulness assumption, we can see $\omega^{T}\mathbf{Y} \nupvDash \mathbf{Z}$ because $\mathbf{E_Y} \nupvDash \mathbf{E_Z}$ (there exists common component $\varepsilon_{L_3}$ for $\varepsilon_{X^{'}_5}$ and $\varepsilon_{X^{'}_6}$), no matter $\omega^\intercal \mathbb{E}[\mathbf{Y}\mathbf{Z}^\intercal] = 0$ or not. 

In Section \ref{Sec:graph_criteria}, we will further investigate graphical implications of GIN in LiNGLaM. For the example given in Figure \ref{fig:simple-example-explain-GIN-constraint}, we have the following observation. $(\{X_4,X_5\},\{X_1,X_2,X_3\})$ satisfies the GIN condition, and  $\{L_1,L_2\}$, the latent common causes 
for $\{X_1,X_2,X_3\}$, d-separate $\{X_1,X_2,X_3\}$ from $\{X_4,X_5\}$. In contrast, $(\{X_3,X_6\},\{X_1,X_2,X_5\})$ violates GIN, and $\{X_1,X_2,X_5\}$ and $\{X_3,X_6\}$ are {\it not} d-separated conditioning on $\{L_1,L_2\}$, the latent common causes of $\{X_1,X_2,X_5\}$.

The following theorem gives mathematical characterizations of the GIN condition, by providing sufficient conditions for when $(\mathbf{Z},\mathbf{Y})$ satisfies the GIN condition. In the next subsection, we give its implication in LiNGLaM; thanks to the constraints implied by the LiNGLaM, one is able to provide sufficient graphical conditions for GIN to hold.

\begin{theorem}\label{Theo-basic}
Suppose that random vectors $\mathbf{L}$, $\mathbf{Y}$, and $\mathbf{Z}$ are related in the following way:
\begin{nospaceflalign}
\mathbf{Y} &= A \mathbf{L} + \mathbf{E}_Y,  \label{eq-X}\\ 
\mathbf{Z} & = B \mathbf{L} + \mathbf{E}_Z. \label{eq-Z}
\end{nospaceflalign}
Denote by $l$ the dimensionality of $\mathbf{L}$.  
Assume $A$ is of full column rank. 
 Then, if 1) $\textrm{Dim}(\mathbf{Y}) > l$, 2) $\mathbf{E}_Y \upvDash \mathbf{L}$, 
    3)  $\mathbf{E}_Y \upvDash \mathbf{E}_Z$,\footnote{Note that we do not assume $\mathbf{E}_Z \upvDash \mathbf{L}$.} and 4) The cross-covariance matrix of $\mathbf{L}$ and $\mathbf{Z}$, $\boldsymbol{\Sigma}_{LZ} = \mathbb{E}[\mathbf{L}\mathbf{Z}^\intercal]$ has rank $l$,  
    then $E_{\mathbf{Y}||\mathbf{Z}} \upvDash \mathbf{Z}$, i.e., $(\mathbf{Z},\mathbf{Y})$ satisfies the GIN condition.
\end{theorem}
\Feng{Theorem 1 gives the mathematical conditions, under which $(\mathbf{Z},\mathbf{Y})$ satisfies the GIN condition. Continue the example in Figure~\ref{fig:simple-example-explain-GIN-constraint}. Let $\mathbf{Z}=\{X_{4},X_{5}\}$ and $\mathbf{Y}=\{X_1,X_2,X_3\}$, and thus $\mathbf{L}=\{L_1,L_2\}$. One then can find the following facts: $\textrm{Dim}(\mathbf{Y})=2 > l$, $\mathbf{E}_Y \upvDash \mathbf{L}$ and $\mathbf{E}_Y \upvDash \mathbf{E}_Z$ according to Eq. 2, and $\boldsymbol{\Sigma}_{LZ} = \mathbb{E}[\mathbf{L}\mathbf{Z}^\intercal]$ has full row rank, i.e., $2$. Therefore, $(\mathbf{Z},\mathbf{Y})$ satisfies the GIN condition.
}
All proofs are given in Supplementary Material.

The following proposition shows that the IN condition can be seen as a special case of the GIN condition with $\mathbf{E}_Z = 0$ (i.e., $\mathbf{Z}$ and $\mathbf{L}$ are linearly deterministically related).
\begin{proposition}\label{Special_case}
Let $\ddot{{Y}} \coloneqq ({Y}, \mathbf{Z})$.  
Then the following statements hold:
\begin{itemize}[noitemsep,topsep=-3pt,leftmargin=30pt]
    \item[1.] $({\mathbf{Z}, \ddot{{Y}}})$ follows the GIN condition if and only if $(\mathbf{Z}, {Y})$ follows it.
    \item[2.] If $(\mathbf{Z}, {Y})$ follows the IN condition, then $(\mathbf{Z}, \ddot{{Y}})$ follows the GIN condition.
\end{itemize}
\end{proposition}
\Feng{Proposition \ref{Special_case} inspires a unified method to handle causal relations between latent variables and those between latent and observed variables. Please see the discussion in Section 6 for more details.}

\vspace{-2mm}
\subsection{Graphical Criteria of GIN in Terms of LiNGLaM}
\vspace{-2mm}
\label{Sec:graph_criteria}
In this section, we investigate graphical implications of the GIN condition in LiNGLaM, which then inspires us to exploit the GIN condition to discover the graph containing latent variables. Specifically, first, the following theorem shows the connection between GIN and the graphical properties of the variables in terms of LiNGLaM. We denote by $L(X_q)$ the set of latent variables that are the parents of $X_q$ and by $L(\mathbf{Y})$ the set of latent variables that are parents of any component of $\mathbf{Y}$. We say variable set $\mathcal{S}_1$ is an {\it exogenous set} relative to  variable set $\mathcal{S}_2$ if and only if 1) $\mathcal{S}_2 \subseteq \mathcal{S}_1$ or 2) for any variable $V$ that is in $\mathcal{S}_2$ but not in $\mathcal{S}_1$, according to the \textcolor{black}{causal graph over $\{V\} \cup \mathcal{S}_1$ and the ancestors of variables in $\{V\} \cup \mathcal{S}_1$, $V$ does not cause any variable in $\mathcal{S}_1$, and the common cause for $V$ and each variable in $\mathcal{S}_1$, if there is any, is also in $\mathcal{S}_1$ (i.e., relative to $\{V\} \cup \mathcal{S}_1$, $V$ does not cause and is not confounded with any variable in $\mathcal{S}_1$).} 
\Feng{For instance, according to the structure in Figure \ref{fig:simple-example-explain-GIN-constraint}, let $\mathcal{S}_1=\{L_1\}$ and $\mathcal{S}_2=\{L_3,L_4\}$, $\mathcal{S}_1$ is an exogenous set relative to $\mathcal{S}_2$. In constrast, if $\mathcal{S}_1=\{L_2, L_3\}$ and $\mathcal{S}_2=\{L_3,L_4\}$, $\mathcal{S}_1$ is not an exogenous set relative to $\mathcal{S}_2$, because 
\textcolor{black}{$L_4$, which is in $\mathcal{S}_2$ but not in $\mathcal{S}_1$, and $L_2$ (as well as $L_3$), which is in $\mathcal{S}_1$, has a common cause, $L_1$, that is not in $\mathcal{S}_1$.}}

\begin{theorem}\label{Theo-2}
Let  $\mathbf{Y}$ and $\mathbf{Z}$ be two disjoint subsets of the observed variables of a LiNGLaM. Assume faithfulness holds for the LiNGLaM. $(\mathbf{Z},\mathbf{Y})$ satisfies the GIN condition if and only if
there exists a $k$-size subset of the latent variables $\mathbf{L}$, $0\leq k \leq \textrm{min}(\textrm{Dim}(\mathbf{Y})-1, \textrm{Dim}(\mathbf{Z}))$, denoted by $\mathcal{S}_{L}^k$, such that 1) $\mathcal{S}_{L}^k$ is an exogenous set relative to $L(\mathbf{Y})$, that 
2) $\mathcal{S}_{L}^k$ d-separates $\mathbf{Y}$ from $\mathbf{Z}$, 
and that 3) the covariance matrix of $\mathcal{S}_{L}^k$ and $\mathbf{Z}$ has rank $k$, and so does that of $\mathcal{S}_{L}^k$ and $\mathbf{Y}$. 
\end{theorem}

Roughly speaking, ${S}_1$ is an  exogenous set relative to  ${S}_2$ if ${S}_1$ contains causally earlier variables (according to the causal order) in or before ${S}_2$. Hence, intuitively,  
the theorem states that  $(\mathbf{Z},\mathbf{Y})$ satisfies the GIN condition when causally earlier common causes of $\mathbf{Y}$ d-separate $\mathbf{Y}$ from $\mathbf{Z}$. We can then see the asymmetry of this condition for $(\mathbf{Z},\mathbf{Y})$ relative to $L(\mathbf{Y})$ and $L(\mathbf{Z})$. For instance, assuming faithfulness, according to the structure in Figure 1, $(\{X_1, X_2\},\{X_3,X_4,X_5\})$ satisfies GIN (with $\mathcal{S}_{L}^2 = \{L_1,L_2\}$), but $(\{X_1, X_6\},\{X_3,X_4,X_5\})$ does not.

Next, we discuss how to identify the group of observed variables that share the same set of latent direct causes; we call such a set of observed variables a \textbf{\textit{causal cluster}}. 
The following theorem formalizes the property of causal clusters and gives a criterion for finding such causal clusters.

\begin{theorem}\label{The-basic-cluster}
	Let $\mathbf{X}$ be the set of all observed variables in a LiNGLaM and $\mathbf{Y}$ be a proper subset of $\mathbf{X}$. If $(\mathbf{X} \setminus \mathbf{Y},\mathbf{Y})$ follows the GIN condition and there is no subset $\tilde{\mathbf{Y}} \subseteq \mathbf{Y}$ such that $(\mathbf{X} \setminus \tilde{\mathbf{Y}},\tilde{\mathbf{Y}})$ follows the GIN condition, then $\mathbf{Y}$ is a causal cluster and $\textrm{Dim}(L(\mathbf{Y}))=\textrm{Dim}(\mathbf{Y})-1$.
\end{theorem}
Consider the example in Figure \ref{fig:simple-example-explain-GIN-constraint}, for $\{X_5,X_6\}$, one can find $(\{X_1,...,X_4,X_7,X_8\},\{X_5,X_6\})$ follows the GIN condition, so $\{X_5,X_6\}$ is a causal cluster and $\textrm{Dim}(L(\{X_5,X_6\}))= \textrm{Dim}(\{X_5,X_6\})-1=1$(i.e., $L_3$).
But, for $\{X_1,X_2,X_5\}$, $(\{X_3,X_4,X_6,X_7,X_8\},\{X_1,X_2,X_5\})$ violates the GIN condition, thus $\{X_1,X_2,X_5\}$ is not a causal cluster.

Furthermore, we discuss how to identify the causal direction between latent variables based on their corresponding children. 
The following theorem shows the asymmetry between the underlying latent variables in terms of the GIN condition.

\begin{theorem}\label{theo: the-basic-causa-direction}
Let $\mathcal{S}_p$ and $\mathcal{S}_q$ be two causal clusters of a LiNGLaM. Assume there is no latent confounder behind $L(\mathcal{S}_p)$ and $L(\mathcal{S}_q)$, and $L(\mathcal{S}_p) \cap L(\mathcal{S}_q)= \emptyset$. Further suppose that $\mathcal{S}_p$ contains $2 \textrm{Dim}(L(\mathcal{S}_p))$ number of variables with $\mathcal{S}_p=\{P_{1},P_{2},...,P_{2 \textrm{Dim}(L(\mathcal{S}_p))}\}$ and that $\mathcal{S}_q$ contains $2 \textrm{Dim}(L(\mathcal{S}_q))$ number of variables with $\mathcal{S}_q=\{Q_{1},Q_{2},...,Q_{2 \textrm{Dim}(L(\mathcal{S}_q))}\}$. Then if $(\{P_{\Feng{\textrm{Dim}(L(\mathcal{S}_p))+1}},...P_{2 \textrm{Dim}(L(\mathcal{S}_p))}\},\{P_{1},....,P_{\textrm{Dim}(L(\mathcal{S}_p))},Q_{1},...Q_{\textrm{Dim}(L(\mathcal{S}_q))}\})$ follows the GIN condition, $L(\mathcal{S}_p) \to L(\mathcal{S}_q)$ holds. 
\end{theorem}
Consider the example in Figure \ref{fig:simple-example-explain-GIN-constraint}.
For two clusters $\{X_1,X_2,X_3,X_4\}$ and $\{X_5,Y_6\}$, where their sets of latent direct causes do not have confounders, one can find $(\{X_3,X_4\},\{X_1,X_2,X_5\})$ follows GIN condition, so $\{L_1,L_2\} \to \{L_3\}$.

\vspace{-2mm}
\section{GIN Condition-Based Algorithm for Estimating LiNGLaM}
\vspace{-2mm}
In this section, we leverage the above theoretical results and propose a recursive algorithm to discover the structural information of LiNGLaM.
The basic idea of the algorithm is that it first finds all causal clusters from the observed data (Step 1), and then it learns the causal order of the latent variables behind these causal clusters (Step 2). \Feng{The completeness of the algorithm is shown in sections 4.1 (Theorem 3 and Proposition 3 for step 1) and 4.2 (Proposition 4 for step 2).}
\vspace{-2mm}
\subsection{Step 1: Finding Causal Clusters}\label{find-clusters}
\vspace{-2mm}
To find causal clusters efficiently, one may start with finding clusters with a single latent variable and merge the overlapping culsters, and then increase the number of allowed latent variables until all variables are put in the clusters. 
We need to consider two practical issues involved in the algorithm. The first is how to find causal clusters and determine how many latent variables they contain, and the second is what clusters should be merged.
Theorem \ref{The-basic-cluster} answers the first question.
Next, for the merge problem, we find that the overlapping clusters can be directly merged into one cluster. This is because the overlapping clusters have the same latent variable as parents in LiNGLaM. The validity of the merging step is guaranteed by Proposition \ref{property_merge}, with the algorithm given in Algorithm \ref{Alg: Cluster}.
\begin{proposition}\label{property_merge}
    Let $\mathcal{S}_1$ and $\mathcal{S}_2$ be two clusters of a LiNGLaM and $\textrm{Dim}(L(\mathcal{S}_1))= \textrm{Dim}(L(\mathcal{S}_2))$. If $\mathcal{S}_1$ and $\mathcal{S}_2$ are overlapping, $\mathcal{S}_1$ and $\mathcal{S}_2$ share the same set of latent variables as parents.
\end{proposition}
\begin{algorithm}[htb]
	\caption{Identifying Causal Clusters}
	\label{alg:one}
	\hspace*{0.02in} {\bf Input:}
	Data set $\mathbf{X} = \{ {X_1},...,{X_m}\}$\\
	\hspace*{0.02in} {\bf Output:}
	Causal cluster set {$\mathbf{\mathcal{S}}$}
	\vspace{-0.3cm}
	\begin{multicols}{2}
		\begin{algorithmic}[1]
		\STATE Initialize $\mathcal{S}= \emptyset$, $Len=1$, and $\mathbf{P}=\mathbf{X}$;
			\REPEAT
			\REPEAT
			\STATE Select a variable subset $\mathcal{P}$ from $\mathbf{P}$ such that $\textrm{Dim}(\mathcal{P})=Len$;
			\IF {$E_{\mathcal{P}||(\mathbf{P} \setminus \mathcal{P})} \upvDash (\mathbf{P} \setminus \mathcal{P})$ holds}
			\STATE $\mathcal{S} =\mathcal{S} \cup \mathcal{P}$;
			\ENDIF
			\UNTIL{{\color{black}{all subsets with length $Len$ in $\mathbf{P}$ have been selected;}}}
			\STATE  Merge all the overlapping sets in $\mathcal{S}$;
			\STATE $\mathbf{P} \leftarrow \mathbf{P} \setminus \mathcal{S}$, and $Len \leftarrow Len+1$;
			\UNTIL{{\color{black}{$\mathbf{P}$ is empty or $\textrm{Dim}(\mathbf{P}) \leq Len$;}}}
			\STATE {\bfseries Return:} $\mathcal{S}$
		\end{algorithmic}
	\end{multicols}
	\vspace{-0.3cm}
	\label{Alg: Cluster}
\end{algorithm}
To test the independence (line 5 in Algorithm \ref{Alg: Cluster}) between two sets of variables, we check for the pairwise independence with the Fisher’s method \citep{fisher1950statistical} instead of testing for the independence between $E_{\mathbf{Y}||\mathbf{Z}}$ and $\mathbf{Z}$ directly. In particular, denote by $p_{k}$, with $k=1,2,...,c$, all resulting $p$-values from pairwise independence tests. We compute the test statistic as $-2\sum_{k=1}^{c}\log{p_k}$, which follows the chi-square distribution with $2c$ degrees of freedom when all the pairs are independent.
\begin{example-set} 
Consider the example in Figure~\ref{fig:simple-example-explain-GIN-constraint}. First, we set $Len=1$ to find the clusters with a single latent variable, i.e., we find $\{X_5,X_6\}$ and $\{X_7,X_8\}$ based on Theorem \ref{The-basic-cluster} (Line 4-9). Then we set $Len=2$ and find the clusters $\{X_1,X_2,X_3,X_4\}$ with two latent variables.
\end{example-set}

\vspace{-2mm}
\subsection{Step 2: Learning the Causal Order of Latent Variables}\label{learn-causal-order}
\vspace{-2mm}
After identifying all clusters, next, we aim to discover the causal order of the set of latent variables of corresponding causal clusters.
As an immediate consequence of Theorem \ref{theo: the-basic-causa-direction}, the root latent variable can be identified by checking the GIN condition, as stated in the following lemma.
\begin{lemma} \label{lemma: Find-Causal-Order-Root-normal}
    Let $\mathcal{S}_r$ be a cluster and $\mathcal{S}_k$, $k \neq r$, be any other cluster of a LiNGLaM. Suppose that $\mathcal{S}_r$ contains $2 \textrm{Dim}(L(\mathcal{S}_r))$ number of variables with $\mathcal{S}_r=\{R_{1},R_{2},...,R_{2 \textrm{Dim}(L(\mathcal{S}_r))}\}$ and that $\mathcal{S}_k$ contains $2 \textrm{Dim}(L(\mathcal{S}_k))$ number of variables with $\mathcal{S}_k=\{K_{1},K_{2},...,K_{2 \textrm{Dim}(L(\mathcal{S}_k))}\}$.
    if $(\{R_{\Feng{\textrm{Dim}(L(\mathcal{S}_r))+1}},...R_{2\textrm{Dim}(L(\mathcal{S}_r))},\{R_{1},....,R_{\textrm{Dim}(L(\mathcal{S}_r))},K_{1},...K_{\textrm{Dim}(L(\mathcal{S}_k))}\})$ follows the GIN condition, then $L(\mathcal{S}_r)$ is a root latent variable set.
\end{lemma}
Now, the key issue is how to use this lemma to recursively discover the ``root variable"\footnote{Note that here we call $L$ a ``root variable" after we have known the variables that causally earlier than $L$.} until the causal order of latent variables is fully determined. Interestingly, we find that in every iteration, we only need to add the children (i.e., the corresponding causal cluster) of the root variable set into the testing set, such that the number of testing latent variables increases when testing the GIN condition in the following steps. Recall the example discussed in Figure 1. For $L_3$, we find that $(\{X_6,\boxed{X_3,X_4}\},\{X_5,X_7,\boxed{X_1,X_2}\})$ satisfies the GIN condition, while for $L_4$, $(\{X_8,\boxed{X_3,X_4}\},\{X_5,X_7,\boxed{X_1,X_2}\})$ violates the GIN condition,\footnote{\Feng{Here, the boxes indicate the elements of the root variable set $\{L_1,L_2\}$}.} which means that $L_3$ is the ``root variable". Intuitively speaking, adding the children of the root variable includes the information of the root variable set and create a new ``root variable", which helps further remove the effect from them. 
Accordingly, we have the following proposition to guarantee the correctness of the above process. The details of the process are given in Algorithm \ref{Alg: Order}.
\begin{proposition} \label{proposition: Find-Causal-Order-Root-normal}
    Suppose that $\{\mathcal{S}_{1},...\mathcal{S}_{i},...,\mathcal{S}_{n}\}$ contains all clusters of the LiNGLaM. Denote $\mathbf{T}=\{L(\mathcal{S}_{1}),...L(\mathcal{S}_{i})\}$ and $\mathbf{R}=\{L(\mathcal{S}_{i+1}),...L(\mathcal{S}_{n})\}$, where all elements in $\mathbf{T}$ are causally earlier than those in $\mathbf{R}$. Let $\mathbf{{\hat Z}}$ contain the elements from the half set of the children of each latent variable set in $\mathbf{T}$, and $\mathbf{{\hat Y}}$ contain the elements from the other half set of the children of each latent variable set in $\mathbf{R}$. Furthermore, 
    Let $L(\mathcal{S}_r)$ be a latent variable set of  $\mathbf{R}$ and $\mathcal{S}_r=\{R_{1},R_{2},...,R_{2 \textrm{Dim}(L(\mathcal{S}_r))}\}$. If for any one of the remaining elements $L(\mathcal{S}_k) \in \mathbf{R}$, with $k \neq r$ and $\mathcal{S}_k=\{K_{1},K_{2},...,K_{2 \textrm{Dim}(L(\mathcal{S}_k))}\}$ such that $(\Feng{\{R_{\textrm{Dim}(L(\mathcal{S}_r))+1},...R_{2\textrm{Dim}(L(\mathcal{S}_r))},\mathbf{{\hat Z}}\},\{R_{1},....,R_{\textrm{Dim}(L(\mathcal{S}_r))},K_{1},...K_{\textrm{Dim}(L(\mathcal{S}_k))},\mathbf{{\hat Y}}\}})$ follows the GIN condition, then $L(\mathcal{S}_r)$ is a root latent variable set in $\mathbf{R}$.
\end{proposition}

\begin{algorithm}[htb]
	\caption{Learning the Causal Order of Latent Variables}
	\label{alg:two}
	\hspace*{0.02in} {\bf Input:}
	Set of causal clusters $\mathcal{S}$\\
	\hspace*{0.02in} {\bf Output:}
	Causal order $\mathcal{K}$
	\vspace{-0.3cm}
	\begin{multicols}{2}
		\begin{algorithmic}[1]
		\STATE Initialize $\mathcal{L}$ with the root variable sets of each cluster, $\mathbf{T}= \emptyset$, and $\mathcal{K}=\emptyset$;
		    \WHILE{$\mathcal{L} \neq \emptyset$}
			\STATE Find the root node $L(\mathcal{S}_{r})$ according to Proposition 4;
			\STATE $\mathcal{L}=\mathcal{L} \setminus L(\mathcal{S}_{r})$;
			\STATE Include $L(\mathcal{S}_{r})$ into the $\mathcal{K}$;
			\STATE $\mathbf{T}=\mathbf{T} \cup \mathcal{S}_{r}$;
			\ENDWHILE
			\STATE {\bfseries Return:} Causal order $\mathcal{K}$
		\end{algorithmic}
	\end{multicols}
	\vspace{-0.3cm}
	\label{Alg: Order}
\end{algorithm}
\begin{example-set}
	Continue to consider the example in Figure~\ref{fig:simple-example-explain-GIN-constraint}. We have found the three causal clusters in step 1, i.e., $\mathcal{S}_{1}=\{X_1,X_2,X_3\}$, $\mathcal{S}_{2}=\{X_5,X_6\}$, and $\mathcal{S}_{3}=\{X_7,X_8\}$.
	Now, we first find that $L(\mathcal{S}_{1})$ is the root variable because $(\{X_3,X_4\},\{X_1,X_2,X_5\})$ and $(\{X_3,X_4\},\{X_1,X_2,X_7\})$ both satisfy the GIN condition (Line 3). Next, we find $L(\mathcal{S}_{2})$ is the ``root variable" because $(\{X_6,X_3,X_4\},\{X5,X7,X_1,X_2\})$ satisfies the GIN condition (Line 3-6). Finally, we return the causal order $\mathcal{K}:L(\mathcal{S}_{1}) \succ L(\mathcal{S}_{2}) \succ L(\mathcal{S}_{3})$. 
\end{example-set}

\vspace{-2mm}
\section{Experimental Results}
\vspace{-2mm}
To show the efficacy of the proposed approach, we applied it to both synthetic and real-world data.
\vspace{-2mm}
\subsection{Synthetic Data}
\vspace{-2mm}
In the following simulation studies, we consider four typical cases: Case 1 \& Case 2 have two latent variables $L_1$ and $L_2$, with $L_1 \to L_2$; Case 3 has three latent variables $L_1$, $L_2$, and $L_3$, with $L_2 \leftarrow L_1 \rightarrow L_3$, and $L_2 \to L_3$; Case 4 has four latent variables $\{L_1,L_2\}$, $L_3$, and $L_4$, with $\{L_1,L_2\} \to L_3$, $\{L_1,L_2\} \to L_4$, and $L_3 \to L_4$.
In all four cases, the data are generated by LiNGLaM and the causal strength $b$ is sampled from a uniform distribution between $[-2,-0.5]\cup[0.5,2]$, noise terms are generated from uniform[-1,1] variables to the fifth power, and the sample size $N = 500, 1000, 2000$.
The details of the graph structures are as follows. [Case 1]: Both $L_1$ and $L_2$ have two pure observed variables, i.e., $L_1 \rightarrow \{X_1,X_2\}$ and $L_2 \rightarrow  \{X_3,X_4\}$. [Case 2]: Add extra edges to the graph in Case 1, such that there exist multiple latent variables. In particular, we add two new variables $\{X_5,X_6\}$, such that $\{L1,L2\} \to \{X_5,X_6\}$, and add the edge $L_1 \to \{X_3, X_4\}$. [Case 3]: Each latent variable has three pure observed variables, i.e., $L_1 \rightarrow  \{X_1,X_2,X_3\}$, $L_2 \rightarrow  \{X_4,X_5,X_6\}$, and $L_3 \rightarrow  \{X_{7},X_{8},X_{9}\}$. [Case 4]: Add extra latent variables and adjust the observed variables in Case 3 such that it becomes the structure in Figure \ref{fig:simple-example-explain-GIN-constraint}.

We compared our algorithm with BPC~\citep{Silva-linearlvModel}, FOFC~\citep{Kummerfeld2016},\footnote{For BPC and FOFC algorithms, we used these implementations in the TETRAD package, which can be downloaded at \url{http://www.phil.cmu.edu/tetrad/}.} and LSTC~\citep{cai2019triad}. 
We measured the estimation accuracy on two tasks: 1) finding the causal clusters, i.e., locating latent variables, and 2) discovering the causal order of latent variables. Note that BPC and FOFC are only applicable to the first task. 

To evaluate the accuracy of the estimated causal cluster, we followed the evaluation metrics from \citet{cai2019triad}. Specifically,
we used \emph{Latent omission}=$\frac{OL}{TL}$, \emph{Latent commission}=$\frac{FL}{TL}$, and \emph{Mismeasurement}=$\frac{MO}{TO}$, where $OL$ is the number of omitted latent variables, $FL$ is the number of falsely detected latent variables, $TL$ is the total number of latent variables in the ground truth graph, $MO$ is the number of falsely observed variables that have at least one incorrectly measured latent, and $TO$ is the number of observed variables in the ground truth graph.
To better evaluate the quality of the estimated causal order, we further used the correct-ordering rate as a metric.
Each experiment was repeated 10 times with randomly generated data and the results were averaged. Here, we used the Hilbert-Schmidt Independence Criterion (HSIC) test~\citep{gretton2008kernel} for the independence test because the data are non-Gaussian.

\begin{center}
\begin{table*}[htp!]
 \vspace{-2mm}
	\small
	\center \caption{results by GIN, LSTC, FOFC, and BPC on learning causal clusters.}
	\label{tab:compare}
	\resizebox{\textwidth}{!}{
	\begin{tabular}{|c|c|c|c|c|c|c|c|c|c|c|c|c|c|}
		\hline  \multicolumn{2}{|c|}{} &\multicolumn{4}{|c|}{\textbf{Latent omission}} & \multicolumn{4}{|c|}{\textbf{Latent commission}} & \multicolumn{4}{|c|}{\textbf{Mismeasurements}}\\
		\hline 
		\multicolumn{2}{|c|}{Algorithm} & GIN & LSTC & FOFC & BPC & GIN & LSTC & FOFC & BPC & GIN & LSTC & FOFC & BPC \\
		\hline 
		 & 500 & 0.00(0) & 0.00(0) & 1.00(10) & 0.50(10) 
		 & 0.00(0) & 0.00(0)& 0.00(0) & 0.00(0) 
		 & 0.00(0) & 0.00(0) & 0.00(0) & 0.00(0) \\
		\cline{2-14}
		{\emph{Case 1}} &1000 &  0.00(0) &0.00(0) & 1.00(10) & 0.50(10) 
		& 0.00(0)&0.00(0) & 0.00(0) & 0.00(0) 
		& 0.00(0) & 0.00(0)& 0.00(0) & 0.00(0) \\
		\cline{2-14}
		&2000 & 0.00(0) & 0.00(0) & 1.00(10) & 0.50(10) 
		& 0.00(0) & 0.00(0)& 0.00(0) & 0.00(0) 
		& 0.00(0) & 0.00(0) & 0.00(0) & 0.00(0) \\
		\hline 
		 & 500 
		 & 0.10(2) & 0.20(4) & 0.9(10) & 0.50(10) 
		 & 0.00(0) & 0.05(1) & 0.00(0) & 0.00(0)
		 & 0.12(2) & 0.12(4) & 0.00(0) & 0.20(10)  \\
		\cline{2-14}
		{\emph{Case 2}} &1000 
		& 0.05(1) & 0.15(3) & 1.00(10) & 0.50(10) 
		& 0.00(0) & 0.00(0) & 0.00(0) & 0.00(0) 
		& 0.04(1) & 0.12(3) & 0.00(0) & 0.20(10)\\
		\cline{2-14} &2000
		& 0.00(0) & 0.00(0) & 1.00(10) & 0.50(10)
		& 0.00(0) & 0.02(2) & 0.00(0) & 0.00(0)
		& 0.00(0) & 0.00(0) & 0.00(0)  & 0.20(10)\\
		\hline 
		 & 500 
		 & 0.20(3) & 0.20(3) & 0.13(9) & 0.10(1) 
		 & 0.00(0) & 0.03(3) & 0.00(0) & 0.00(0) 
		 & 0.19(3) & 0.17(3) & 0.00(0) & 0.00(0)\\
		\cline{2-14}
		{\emph{Case 3}} &1000 
		& 0.06(2) & 0.13(2) & 0.16(10) & 0.00(0) 
		& 0.00(0) & 0.00(0) & 0.00(0) & 0.00(0) 
		& 0.06(2) & 0.00(0) &0.00(0)  & 0.00(0)\\
		\cline{2-14}&2000 
		& 0.00(0) & 0.00(0) & 0.50(10) & 0.00(0)
		& 0.00(0) & 0.00(0) & 0.00(0) & 0.00(0)
		& 0.00(0) & 0.00(0) & 0.00(0) & 0.00(0)\\
		\hline 
		 & 500 
		 &0.13(4) &0.40(6) & 0.90(10) & 0.63(10)
		 &0.00(0) &0.23(5) & 0.00(0) & 0.00(0)
		 &0.04(2) &0.15(6) & 0.02(2) & 0.06(4)\\
		\cline{2-14}
		{\emph{Case 4}} &1000
	     &0.10(3) &0.26(6) & 0.93(10) & 0.66(10)
		 &0.00(0) &0.00(0) & 0.00(0) & 0.00(0)
		 &0.05(3) &0.11(2) & 0.01(1) & 0.02(2)\\
		\cline{2-14}
		&2000 
		 &0.03(1) &0.32(6) & 1.00(10) & 0.70(10)
		 &0.00(0) &0.00(0) & 0.00(0) & 0.00(0)
		 &0.04(1) &0.11(3) & 0.00(10) & 0.00(0)\\    
		\hline 
	\end{tabular}}
	\begin{tablenotes}
		\item Note: The number in parentheses indicates the number of occurrences that the current algorithm {\it cannot} correctly solve the problem. 
	\end{tablenotes}
	\vspace{-2mm}
\end{table*}
\end{center}

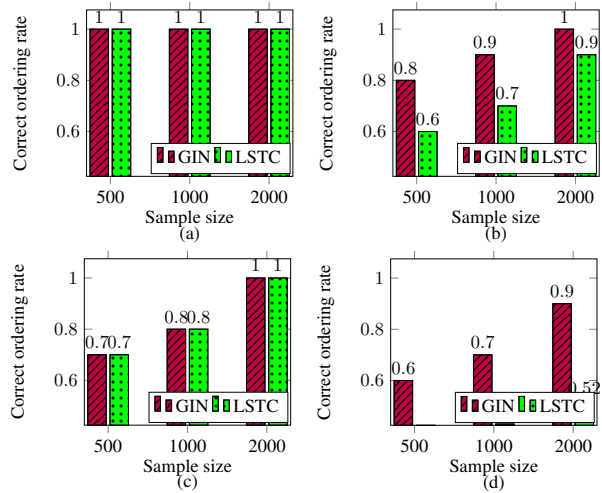
\begin{figure}[htp!] 
   \setlength{\abovecaptionskip}{0pt}
	\setlength{\belowcaptionskip}{0pt}
	\label{fig:exp}
	\begin{center}
		\begin{tikzpicture}[scale=0.7]
		\begin{axis}[
		title=(a),
		title style={at={(0.5,-0.5)}},
		width=5.5cm,
		cycle list name=mark list,
		ybar,
		enlargelimits=0.15,
        ymin=0.5,ymax=1,
		legend style={at={(0.65,0.2)}, anchor=north,legend columns=-1},
		ylabel={Correct ordering rate},
		xlabel={Sample size},
		symbolic x coords={500,1000,2000},
		xtick=data,
		nodes near coords,
		nodes near coords align={vertical},
		]
		\addplot [ fill=purple,postaction={pattern=north east lines}] coordinates {(500,1) (1000,1) (2000,1)};
		\addplot [fill=green,postaction={pattern=dots}] coordinates {(500,1) (1000,1) (2000,1)};
		\legend{GIN,LSTC}
		\end{axis}
		\end{tikzpicture}~~
		\begin{tikzpicture}[scale=0.7]
		\begin{axis}[
		title=(b),
		title style={at={(0.5,-0.5)}},
		width=5.5cm,
		cycle list name=mark list,
		ybar,
		enlargelimits=0.15,
        ymin=0.5,ymax=1,
		legend style={at={(0.65,0.2)}, anchor=north,legend columns=-1},
		ylabel={Correct ordering rate},
		xlabel={Sample size},
		symbolic x coords={500,1000,2000},
		xtick=data,
		nodes near coords,
		nodes near coords align={vertical},
		]
		\addplot [ fill=purple,postaction={pattern=north east lines}] coordinates {(500,0.8) (1000,0.9) (2000,1)};
		\addplot [fill=green,postaction={pattern=dots}] coordinates {(500,0.6) (1000,0.7) (2000,0.9)};
		\legend{GIN,LSTC}
		\end{axis}
		\end{tikzpicture}\\
		\begin{tikzpicture}[scale=0.7]
		\begin{axis}[
		title=(c),
		title style={at={(0.5,-0.5)}},
		width=5.5cm,
		cycle list name=mark list,
		ybar,
		enlargelimits=0.15,
        ymin=0.5,ymax=1,
		legend style={at={(0.65,0.2)}, anchor=north,legend columns=-1},
		ylabel={Correct ordering rate},
		xlabel={Sample size},
		symbolic x coords={500,1000,2000},
		xtick=data,
		nodes near coords,
		nodes near coords align={vertical},
		]
		\addplot [ fill=purple,postaction={pattern=north east lines}] coordinates {(500,0.7) (1000,0.8) (2000,1)};
		\addplot [fill=green,postaction={pattern=dots}] coordinates {(500,0.7) (1000,0.8) (2000,1)};
		\legend{GIN,LSTC}
		\end{axis}
		\end{tikzpicture}~~
		\begin{tikzpicture}[scale=0.7]
		\begin{axis}[
		title=(d),
		title style={at={(0.5,-0.5)}},
		width=5.5cm,
		cycle list name=mark list,
		ybar,
		enlargelimits=0.15,
        ymin=0.5,ymax=1,
		legend style={at={(0.65,0.2)}, anchor=north,legend columns=-1},
		ylabel={Correct ordering rate},
		xlabel={Sample size},
		symbolic x coords={500,1000,2000},
		xtick=data,
		nodes near coords,
		nodes near coords align={vertical},
		]
		\addplot [ fill=purple,postaction={pattern=north east lines}] coordinates {(500,0.6) (1000,0.7) (2000,0.9)};
		\addplot [fill=green,postaction={pattern=dots}] coordinates {(500,0.37) (1000,0.43) (2000,0.52)};
		\legend{GIN,LSTC}
		\end{axis}
		\end{tikzpicture}
		\caption{ (a-d) Accuracy of the estimated causal order with GIN (purple), and LSTC (green) for Cases 1-4.}
		\vspace{0cm}
		\label{fig:F1-scores}
	\end{center}
\end{figure}
As shown in Table \ref{tab:compare}, our algorithm, GIN, achieves the best performance (the lowest errors) on \Feng{almost all cases of the structures. We noticed that although the Mismeasurements of GIN are higher than LSTC in Case 3 when the sample size is small (N=500), the Latent commission of GIN are lower than LSTC. 
}
The BPC and FOFC algorithms (with distribution-free tests) do not perform well, which implies that the rank constraints on covariance matrix is not enough to recover more latent structures.
Interestingly, although the LSTC algorithm has low errors of the Latent omission in Case 2 (it may be because the structure in Case 2 can be transformed into equivalent pure structures~\citep{cai2019triad}), it can not tell us the number of latent variables behind observed variables. Moreover, LSTC fails to recover Case 4 because of the multiple latent variables.
The above results demonstrate a clear advantage of our method over the comparisons. 

Considering that BPC and FOFC algorithms can not discover the causal directions of latent variables, we only reported the results of LSTC algorithm and our algorithm on causal order learning in Figure~\ref{fig:F1-scores}.
As shown in Figure~\ref{fig:F1-scores}, the accuracy of the identified causal ordering of our method gradually increases to 1 with the sample size in all the four cases. LSTC can not handle Case 2 \& 4. These findings illustrate that our algorithm can discover the correct causal order.
\vspace{-3mm}
\subsection{Real-World Data}
\vspace{-3mm}
\begin{wrapfigure}{r}{0.53\textwidth} 
   \setlength{\abovecaptionskip}{2pt}
	\setlength{\belowcaptionskip}{-6pt}
	\small
    \centering
    \vspace{-0.3cm}
    \begin{tabular}{|c|c|}
\hline
Causal Clusters & Observed variables \\ \hline
$\mathcal{S}_{1}$ (1) & $RC_{1}$, $RC_{2}$, $WO_{1}$, $WO_{2}$,  \\ & $DM_{1}$, $DM_{2}$\\\hline
$\mathcal{S}_{2}$ (1) & $CC_{1}$, $CC_{2}$,$CC_{3}$,$CC_{4}$\\\hline
$\mathcal{S}_{3}$ (1) & $PS_{1}$, $PS_{2}$\\\hline
$\mathcal{S}_{4}$ (1) & $ELC_{1}$, $ELC_{2}$,$ELC_{3}$,$ELC_{4}$,  \\ &$ELC_{5}$\\\hline
$\mathcal{S}_{5}$ (2) & $SE_{1}$, $SE_{2}$, $SE_{3}$, $EE_{1}$,  \\ & $EE_{2}$, $EE_{3}$,  $DP_{1}$, $PA_{3}$\\ \hline
$\mathcal{S}_{6}$ (3) & $DP_{2}$, $PA_{1}$, $PA_{2}$ \\
\hline
    \end{tabular}
    \caption{The output of Algorithm 1 in the teacher's burnout study.}
    \vspace{-0.4cm}
    \label{tab:causal-clusters}
\end{wrapfigure}
Barbara Byrne conducted a study to investigate the impact of organizational (role ambiguity, role conflict, classroom climate, and superior support, etc.) and personality (self-esteem, external locus of control) on three facets of burnout in full-time elementary teachers \citep{byrne2010structural}. 
We applied our algorithm to this data set, with 28 observed variables in total.

In the implementation, the kernel width in the HSIC test is set to 0.05. We first applied Algorithm 1 and received six causal clusters, including one cluster with $2$ latent variables and one cluster with 3 latent variables. The results were given in Figure \ref{tab:causal-clusters}. Next, we applied Algorithm 2 and got the final causal order (from root to leaf): $L(\mathcal{S}_{1}) \succ L(\mathcal{S}_{2}) \succ L(\mathcal{S}_{3}) \succ L(\mathcal{S}_{5}) \succ L(\mathcal{S}_{4}) \succ L(\mathcal{S}_{6})$. Specifically, we had the following findings. 1. The identified clusters are similar to the domain knowledge, e.g, $\mathcal{S}_{2}$ represents the classroom climate, $\mathcal{S}_{3}$ represents the peer support, $\mathcal{S}_{4}$ represents the external locus of control, et al. 2. The learned causal order is similar to Byrne's conclusion, e.g., personal accomplishment ($L(\mathcal{S}_{6})$) are caused by other latent factors. In addition, role conflict 
and decision making ($L(\mathcal{S}_{1})$), classroom climate ($L(\mathcal{S}_{2})$), and peer support ($L(\mathcal{S}_{3})$) cause burnout (including emotional exhaustion, depersonalization, and personal accomplishment ($L(\mathcal{S}_{5})$ and $L(\mathcal{S}_{6})$)).

\vspace{-3mm}
\section{Discussion and Further Work}
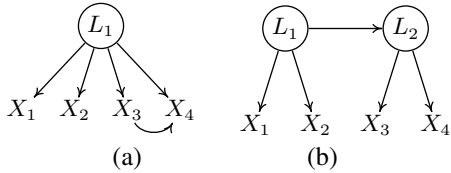
\begin{figure}[htp!]
	\begin{center}
        \begin{tikzpicture}[scale=1.4, line width=0.5pt, inner sep=0.2mm, shorten >=.1pt, shorten <=.1pt]
		\draw (0.75, 0) node(L1) [circle, draw] {{\footnotesize\,$L_1$\,}};
		\draw (0, -0.8) node(X1) [] {{\footnotesize\,${X}_{1}$\,}};
		\draw (0.5, -0.8) node(X2) [] {{\footnotesize\,$X_2$\,}};
		\draw (1, -0.8) node(X3) [] {{\footnotesize\,$X_3$\,}};
		\draw (1.5, -0.8) node(X4) [] {{\footnotesize\,${X}_{4}$\,}};
		\draw[-arcsq] (L1) -- (X1) node[pos=0.5,sloped,above] {};
		\draw[-arcsq] (L1) -- (X2) node[pos=0.5,sloped,above] {};
		\draw[-arcsq] (L1) -- (X3) node[pos=0.5,sloped,above] {};
		\draw[-arcsq] (L1) -- (X4) node[pos=0.5,sloped,above] {}; 
		\draw [-arcsq] (X3) edge[bend right=60] (X4);
		\end{tikzpicture}~~~~
		\begin{tikzpicture}[scale=1.6, line width=0.5pt, inner sep=0.2mm, shorten >=.1pt, shorten <=.1pt]
		\draw (0.5, 0) node(L1) [circle, draw] {{\footnotesize\,$L_1$\,}};
		\draw (1.5, 0) node(L2) [circle, draw] {{\footnotesize\,$L_2$\,}};
		\draw (0.25, -0.8) node(X1) [] {{\footnotesize\,${X}_{1}$\,}};
		\draw (0.75, -0.8) node(X2) [] {{\footnotesize\,$X_2$\,}};
		\draw (1.25, -0.8) node(X3) [] {{\footnotesize\,$X_3$\,}};
		\draw (1.75, -0.8) node(X4) [] {{\footnotesize\,${X}_{4}$\,}};
		\draw[-arcsq] (L1) -- (X1) node[pos=0.5,sloped,above] {};
		\draw[-arcsq] (L1) -- (X2) node[pos=0.5,sloped,above] {};
		\draw[-arcsq] (L2) -- (X3) node[pos=0.5,sloped,above] {};
		\draw[-arcsq] (L2) -- (X4) node[pos=0.5,sloped,above] {}; 
		\draw[-arcsq] (L1) -- (L2)node[pos=0.5,sloped,above]{};
		\end{tikzpicture}\\
		(a)~~~~~~~~~~~~~~~~~~~~~~~~~(b) \vspace{-0.1cm}
		\caption{Two structures that are distinguishable by the GIN condition, while (a) has an edge between observed variables $X_3$ and $X_4$. } \vspace{-0.2cm}
		\label{fig:discussion} 
	\end{center}
\end{figure}

\vspace{-3mm}
The preceding sections presented how to use GIN conditions to locate the latent variables and identify their causal structure in the LiNGLaM. In this procedure we examine whether the ordered pair of two disjoint subsets of the observed variables satisfies GIN. As shown in Proposition \ref{Special_case}, the GIN condition actually contains IN as a special case, in which the two subsets of variables have overlapping variables. For instance, suppose we have only two variables with $X_1 \rightarrow X_2$. Then $(X_1,X_2)$ satisties IN, and $(X_1, (X_2, X_1))$ satisfies GIN. As a consequence, 
interestingly, even if we allow edges between observed variables in the LiNGLaM, the GIN condition may also be used to identify them, together with their connections. For instance, in Figure \ref{fig:discussion}(a), $(\{X_{1},X_{3}\},\{X_{2},X_{3},X_{4}\})$ satisfies the GIN condition while $(\{X_{1},X_{4}\},\{X_{2},X_{3},X_{4}\})$ violates the 
GIN condition, which means that there is an edge between $X_{3}$ and $X_{4}$ and $X_{3} \to X_{4}$. 
In contrast, with the GIN condition on pairs of disjoint subsets of variables, one cannot distinguish between structures (a) and (b). Developing an efficient algorithm that is able to recover the LiNGLaM with directed edges between observed variables in a principled way is part of our future work. Furthermore, in this paper we focus on discovery of the structure of the LiNGLaM, more specifically, the locations of the latent variables and their causal order; as future work, we will also show the (partial) identifiability of the causal coefficients in the model and develop an estimation method for them, to produce a fully specified estimated LiNGLaM (further with edges between observed variables).

\vspace{-3mm}
\section{Conclusion}
\vspace{-3mm}
We proposed a Generalized Independent Noise (GIN) condition for estimating a particular type of linear non-Gaussian latent variable causal model, which includes the Independent Noise (IN) condition as a special case. We showed the graphical implications of the GIN condition, based on which we proposed a recursive learning algorithm to locate latent causal variables and identify their causal structure.
Experimental results on simulation data and real data further verified the usefulness of our algorithm. 
Future research along this line includes allowing casual edges between observed variables and allowing nonlinear causal relationships. 

\vspace{-1mm}
\subsubsection*{Acknowledgments}
\vspace{-1mm}
This research was supported in part by Natural Science Foundation of China (61876043) and Science and Technology Planning Project of Guangzhou(201902010058) and Outstanding Young Scientific Research Talents International Cultivation Project Fund of Department of Education of Guangdong Province(40190001). KZ would like to acknowledge
the support by the United States Air Force under Contract
No. FA8650-17-C-7715.
We appreciate the comments from Peter Spirtes and anonymous
reviewers, which greatly helped to improve the paper.


\appendix

\section*{Supplementary Material}
The supplementary material contains
\begin{itemize}
    \item Proof of Proposition 1;
    \item Proof of Theorem 1;
    \item Proof of Proposition 2;
    \item Proof of (and remark on) Theorem 2;
    \item Proof of Theorem 3;
    \item Proof of Theorem 4;
    \item Proof of Proposition 3;
    \item Proof of Lemma 1;
    \item Proof of Proposition 4;
    \item More experimental results of Synthetic data;
    \item More details of Real-Word data.
\end{itemize}

\section{Proofs and Illustrations}
We first give an important theorem, which will be used in the proof.

\noindent{\rm{\textbf{Darmois-Skitovitch Theorem ~\citep{Kagan-1973book}}}}
Define two random variables ${X_1}$ and ${X_2}$ as linear combinations of independent random variables ${e_i} (i = 1,...,p)$:
\begin{flalign}
{X_1} = \sum\limits_{i = 1}^p {{\alpha _i}} {e_i}, \quad \quad {X_2} = \sum\limits_{i = 1}^q {{\beta _i}} {e_i}.
\end{flalign}
Then, if ${X_1}$ and ${X_2}$ are independent, all variables ${e_j}$ for which ${\alpha _j}{\beta _j} \ne 0$ are Gaussian. 
In other words, if there exists a non-Gaussian ${e_j}$ for which ${\alpha _j}{\beta _j} \ne 0$, ${X_1}$ and ${X_2}$ are dependent.
\subsection{Proof of Proposition 1}
\begin{repproposition}{pro-extension-IN}
Suppose all considered variables follow the linear non-Gaussian acyclic causal model. Let $\mathbf{Z}$ be a subset of those variables and $Y$ be a single variable among them. Then the following statements are equivalent.
\begin{itemize}
    \item[(A)] 1) All variables in $\mathbf{Z}$ are causally earlier than $Y$,  and 2) There is no common cause for each variable in $\mathbf{Z}$ and $Y$ that is not in $\mathbf{Z}$.
    \item[(B)] $(\mathbf{Z},Y)$ satisfies the IN condition.
\end{itemize}
\end{repproposition}
The proof is straightforward, based on the assumption of linear, non-Gaussian acyclic causal models.
\subsection{Proof of Theorem 1}
\begin{reptheorem}{Theo-basic}
Suppose that random vectors $\mathbf{L}$, $\mathbf{Y}$, and $\mathbf{Z}$ are related in the following way:
\begin{flalign}
\mathbf{Y} &= A \mathbf{L} + \mathbf{E}_Y, \\ 
\mathbf{Z} & = B \mathbf{L} + \mathbf{E}_Z.
\end{flalign}
Denote by $l$ the dimensionality of $\mathbf{L}$.  
Assume $A$ is of full column rank. 
 Then, if 1) $\textrm{Dim}(\mathbf{Y}) > l$, 2) $\mathbf{E}_Y \upvDash \mathbf{L}$, 
    3)  $\mathbf{E}_Y \upvDash \mathbf{E}_Z$,\footnote{Note that we do not assume $\mathbf{E}_Z \upvDash \mathbf{L}$.} and 4) The cross-covariance matrix of $\mathbf{L}$ and $\mathbf{Z}$, $\boldsymbol{\Sigma}_{LZ} = \mathbb{E}[\mathbf{L}\mathbf{Z}^\intercal]$ has rank $l$,  
    then $E_{\mathbf{Y}||\mathbf{Z}} \upvDash \mathbf{Z}$, i.e., $(\mathbf{Z},\mathbf{Y})$ satisfies the GIN condition.
\end{reptheorem}

\begin{proof}
Without loss of generality, assume that each component of $\mathbf{L}$ has a zero mean, and that both $\mathbf{E}_Y$ and $\mathbf{E}_Z$ are zero-mean. 
If we can find a non-zero vector $\omega$ such that $\omega^\intercal A = 0$, then $\omega^\intercal \mathbf{Y} =\omega^\intercal A\mathbf{L} + \omega^\intercal \mathbf{E}_Y = \omega^\intercal \mathbf{E}_Y$, which will be independent from $\mathbf{Z}$ in light of conditions 2) and 3), i.e., the GIN condition for $\mathbf{Y}$ given $\mathbf{Z}$ holds true.

We now construct the vector $\omega$. If conditions 2) and 3) hold, we have
$\mathbb{E}[\mathbf{Y}\mathbf{Z}^\intercal] =  A \boldsymbol{\Sigma}_{LZ}$, which is determined by $(\mathbf{Y},\mathbf{Z})$.
We now show that under conditions 4), for any non-zero vector $\omega$, $\omega^\intercal A = 0$ if and only if $\omega^\intercal A \boldsymbol{\Sigma}_\mathbf{LZ} = 0$ or equivalently $\omega^\intercal \mathbb{E}[\mathbf{Y}\mathbf{Z}^\intercal] = 0$ and that such a vector $\omega$ exists.

Suppose $\omega^\intercal A = 0$, it is trivial to see $\omega^\intercal A \boldsymbol{\Sigma}_{LZ} = 0$. Notice that condition 4) implies that $\textrm{rank}(A \boldsymbol{\Sigma}_{LZ}) \leq l$ because $\textrm{rank}(A \boldsymbol{\Sigma}_{LZ}) \leq \textrm{min}(\textrm{rank}(A), \textrm{rank}(\boldsymbol{\Sigma}_{LZ}))$ and $\textrm{rank}(A) = l$. Further according to Sylvester Rank Inequality, we have $\textrm{rank}(A \boldsymbol{\Sigma}_{LZ}) \geq \textrm{rank}(A) + \textrm{rank}( \boldsymbol{\Sigma}_{LZ}) - l = l$. Therefore, $\textrm{rank}(A \boldsymbol{\Sigma}_{LZ}) = l$. Because of condition 1), there must exists a non-zero vector $\omega$, determined by $(\mathbf{Y,Z})$, such that $\omega^\intercal \mathbb{E}[\mathbf{YZ}^\intercal] = \omega^\intercal A \boldsymbol{\Sigma}_{LZ}  = 0$. Moreover, this equality implies $\omega^\intercal A = 0$ because $\boldsymbol{\Sigma}_{LZ}$ has $l$ rows and has rank $l$. With this $\omega$, we have  $E_{\mathbf{Y}||\mathbf{Z}} = \omega^\intercal \mathbf{E}_Y$ and is independent from $\mathbf{Z}$. Thus the theorem holds.
\end{proof}

\subsection{Proof of Proposition 2}

\begin{repproposition}{Special_case}
Let $\ddot{{Y}} \coloneqq ({Y}, \mathbf{Z})$.  
Then the following statements hold:
\begin{itemize}[noitemsep,topsep=-3pt,leftmargin=30pt]
    \item[1.] $(\mathbf{Z}, \ddot{{Y}})$ follows the GIN condition if and only if $(\mathbf{Z}, {Y})$ follows it.
    \item[2.] If $(\mathbf{Z}, {Y})$ follows the IN condition, then $(\mathbf{Z}, \ddot{{Y}})$ follows the GIN condition.
\end{itemize}
\end{repproposition}

\begin{proof}
For Statement 1, we first show that $(\mathbf{Z}, \ddot{{Y}})$ follows the GIN condition implies that $(\mathbf{Z}, {Y})$ follows the GIN condition.
If $(\mathbf{Z}, \ddot{{Y}})$ follows the GIN condition, then there must exist a non-zero vector $\ddot{\omega}$ so that $\ddot{\omega}^\intercal \mathbb{E}[\ddot{Y}\mathbf{Z}^\intercal] =0$. This equality implies 
\begin{align}
    \ddot{\omega}^\intercal \mathbb{E}[\left[\begin{matrix}
{Y}\\
{\mathbf{Z}}
\end{matrix}\right]\mathbf{Z}^\intercal]=\ddot{\omega}^\intercal \left[\begin{matrix}
{\mathbb{E}[Y\mathbf{Z}^\intercal]}\\
{\mathbb{E}[\mathbf{Z}\mathbf{Z}^\intercal}]
\end{matrix}\right]  =\mathbf{0}.
\end{align}
Because $\mathbb{E}[\mathbf{ZZ}^\intercal]$ is  non-singular, we further have
$$\ddot{\omega}^\intercal \left[\begin{matrix}
{\mathbb{E}[Y\mathbf{Z}^\intercal]\mathbb{E}^{-1}[\mathbf{Z}\mathbf{Z}^{\intercal}]}\\
{\mathbf{I}}
\end{matrix}\right] = \mathbf{0}.$$
Let $\omega$ be the first \textit{Dim}$(Y)$ dimensions of $\ddot{\omega}$. Then we have 
${\omega}^\intercal \mathbb{E}[{Y}\mathbf{Z}^\intercal]\mathbb{E}^{-1}[\mathbf{Z}\mathbf{Z}^{\intercal}] =\mathbf{0}$, and thus ${\omega}^\intercal \mathbb{E}[{Y}\mathbf{Z}^\intercal]=\mathbf{0}$. 
Furthermore, based on the definition of the GIN condition, we have that
$E_{\ddot{Y}||\mathbf{Z}} = \ddot{\omega}^\intercal \ddot{Y}$ is independent from $\mathbf{Z}$. It is easy to see that $E_{{Y}||\mathbf{Z}} = {\omega}^\intercal Y$ is independent from $\mathbf{Z}$. Thus, $(\mathbf{Z}, {Y})$ follows the GIN condition.

Next, we show that $(\mathbf{Z}, {Y})$ follows the GIN condition implies that $(\mathbf{Z}, \ddot{{Y}})$ follows the GIN condition.
If $(\mathbf{Z}, {Y})$ follows the GIN condition, we have 
\begin{align}
    {\omega}^\intercal \mathbb{E}[{Y}\mathbf{Z}^\intercal]=\mathbf{0}
\end{align}
Let $\ddot{\omega}=[\omega^\intercal \;, \mathbf{0}^\intercal]^\intercal$. We have
\begin{align}
    \ddot{{\omega}}^\intercal \mathbb{E}[{\ddot{Y}}\mathbf{Z}^\intercal]=[\omega^\intercal \;, \mathbf{0}^\intercal] \mathbb{E}[\left[\begin{matrix}
{Y}\\
{\mathbf{Z}}
\end{matrix}\right]\mathbf{Z}^\intercal]={\omega}^\intercal \mathbb{E}[{Y}\mathbf{Z}^\intercal]=\mathbf{0}.
\end{align}
Furthermore, we have $\ddot{{\omega}}^\intercal \ddot{Y}=[\omega^\intercal \;, 0^\intercal] \left[\begin{matrix}
{Y}\\
{\mathbf{Z}}
\end{matrix}\right] =\omega ^\intercal{Y}$.
Based on the definition of GIN, $E_{{Y}||\mathbf{Z}} = {\omega}^\intercal Y$ is independent from $\mathbf{Z}$.
That is to say, $\ddot{\omega}^\intercal \mathbf{E}_{\ddot{Y}}$ is independent from $\mathbf{Z}$. Thus, $(\mathbf{Z}, \ddot{Y})$ follows the GIN condition.

For Statement 2, If $(\mathbf{Z}, {Y})$ follows the IN condition, we have
\begin{align}\label{eq-pro1-2-w}
    \tilde{\omega} = \mathbb{E}[Y \mathbf{Z}^\intercal]\mathbb{E}^{-1}[\mathbf{Z} \mathbf{Z}^\intercal].
\end{align}
Let $\ddot{\omega}=[1^\intercal \; ,-\tilde{\omega}^\intercal]^\intercal$, we get
\begin{align}\label{eq-pro1-2-wEYZ}
    \ddot{{\omega}}^\intercal \mathbb{E}[{\ddot{Y}}\mathbf{Z}^\intercal]=[1^\intercal \; ,-\tilde{\omega}^\intercal] \mathbb{E}[\left[\begin{matrix}
{Y}\\
{\mathbf{Z}}
\end{matrix}\right]\mathbf{Z}^\intercal]=[1^\intercal \; ,-\tilde{\omega}^\intercal] \left[\begin{matrix}
{\mathbb{E}[Y\mathbf{Z}^\intercal]}\\
{\mathbb{E}[\mathbf{Z}\mathbf{Z}^\intercal}]
\end{matrix}\right] ={\mathbb{E}[Y\mathbf{Z}^\intercal]}-\tilde{\omega}{\mathbb{E}[\mathbf{Z}\mathbf{Z}^\intercal}].
\end{align}
From Equations \ref{eq-pro1-2-w} and \ref{eq-pro1-2-wEYZ}, we have $\ddot{{\omega}}^\intercal \mathbb{E}[{\ddot{Y}}\mathbf{Z}^\intercal]=0$. That is to say, $\ddot{{\omega}}$ satisfies $\ddot{{\omega}}^{\intercal}\mathbb{E}[{\ddot{Y}}\mathbf{Z}^\intercal]=0$ and that $\ddot{{\omega}}^{\intercal} \neq \mathbf{0}$.

Now, we show that $\ddot{{\omega}}^{\intercal}\ddot{Y}$ is independent from $\mathbf{Z}$. We know that $Y-\tilde{\omega}^{\intercal}\mathbf{Z}$ is independent from $\mathbf{Z}$ based on the definition of the IN condition.
It is easy to see that $\ddot{{\omega}}^{\intercal}\ddot{Y}=[1^\intercal \; ,-\tilde{\omega}^\intercal]\left[\begin{matrix}
{Y}\\
{\mathbf{Z}}
\end{matrix}\right]=Y-\tilde{\omega}^{\intercal}\mathbf{Z}$ is independent from $\mathbf{Z}$. Therefore, $(\mathbf{Z}, \ddot{{Y}})$ follows the GIN condition.

\end{proof}

\subsection{Proof of and Remark on Theorem 2}
\begin{reptheorem}{Theo-2}
Let  $\mathbf{Y}$ and $\mathbf{Z}$ be two disjoint sets of observed variables of a LiNGLaM. Assume faithfulness holds for the LiNGLaM. 
$(\mathbf{Z},\mathbf{Y})$ satisfies the GIN condition if and only if
there exists a $k$-size subset of the latent variables $\mathbf{L}$, $0\leq k \leq \textrm{min}(Dim(\mathbf{Y})-1, Dim(\mathbf{Z}))$, denoted by $\mathcal{S}_{L}^k$, such that 1) $\mathcal{S}_{L}^k$ is an exogenous set relative to $L(\mathbf{Y})$, that 
2) $\mathcal{S}_{L}^k$ d-separates $\mathbf{Y}$ from $\mathbf{Z}$, 
and that 3) the covariance matrix of $\mathcal{S}_{L}^k$ and $\mathbf{Z}$ has rank $k$, and so does that of $\mathcal{S}_{L}^k$ and $\mathbf{Y}$. 
\end{reptheorem}
\begin{proof}
The ``if" part: First suppose that there exists such a subset of the latent variables,  $\mathcal{S}_{L}^k$, that satisfies the three conditions. Because of condition 1), i.e., that $\mathcal{S}_{L}^k$ is an exogenous set relative to $L(\mathbf{Y})$ and because according to the LiNGLaM, each ${Y}_i$ is a linear function of $L(Y_i)$ plus independent noise, we know that $\mathcal{S}_{L}^k$ is also an exogenous set relative to $\mathbf{Y}$. Hence,  we know that each component of $\mathbf{Y}$ can be written as a linear function of $\mathcal{S}_{L}^k$ and some independent error (which is independent from $\mathcal{S}_{L}^k$). By a slight abuse of notation, here we use $\mathcal{S}_{L}^k$ also to denote the vector of the variables in $\mathcal{S}_{L}^k$. Then we have 
\begin{equation} \label{eq_t1}
    \mathbf{Y} = A \mathcal{S}_{L}^k + \mathbf{E}'_Y,
\end{equation}
where $A$ is an appropriate linear transformation, $\mathbf{E}'_Y$ is independent from $\mathcal{S}_{L}^k$, but its components are not necessarily independent from each other. In fact, according to the LiNGLaM, each observed or hidden variable is a linear combination of the underlying noise terms $\varepsilon_i$. In equation (\ref{eq_t1}), $\mathcal{S}_{L}^k$ and $\mathbf{E}'_Y$ are linear combinations of disjoint sets of the noise terms $\varepsilon_i$, implied by the directed acyclic structure over all observed and hidden variables.

Let us then write $\mathbf{Z}$ as linear combinations of the noise terms. We then show that because of condition 2), i.e., that $\mathcal{S}_{L}^k$ d-separates $\mathbf{Y}$ from $\mathbf{Z}$, if any noise term $\varepsilon_i$ is present in $\mathbf{E}'_Y$, it will not be among the noise terms in the expression of  $\mathbf{Z}$. Otherwise, if $Z_j$ also involves $\varepsilon_i$, then the direct effect of $\varepsilon_i$, among all observed or hidden variables, is a common cause of $Z_j$ and some component of $\mathbf{Y}$. This path between $Z_j$ and that component of $\mathbf{Y}$, however, cannot be d-separated by $\mathcal{S}_{L}^k$ because no component of $\mathcal{S}_{L}^k$ is on the path, as implied by the fact that 
when $\mathcal{S}_{L}^k$ is written as a linear combination of the underlying noise terms, $\varepsilon_i$ is not among them. Consequently, any noise term in $\mathbf{E}'_Y$ will not contribute to $\mathcal{S}_{L}^k$ or  $\mathbf{Z}$. Hence, we can express $\mathbf{Z}$ as
\begin{equation} \label{eq_t2}
    \mathbf{Z} = B \mathcal{S}_{L}^k + \mathbf{E}_Z',
\end{equation}
where $\mathbf{E}_Z'$, which is determined by $\mathcal{S}_{L}^k$ and $\mathbf{Z}$, is independent from $\mathbf{E}_Y'$. Further considering condition on the dimensionality of $\mathcal{S}_{L}^k$ and condition 3), one can see that the assumptions in Theorem 1 are satisfied. Therefore, $(\mathbf{Z},\mathbf{Y})$ satisfies the GIN condition.

The ``only-if" part: Then we suppose $(\mathbf{Z,Y})$ satisfies GIN (while with the same $\mathbf{Z}$, no proper subset of $\mathbf{Y}$ does). Consider all sets $\mathcal{S}_{L}^k$ that are exogenous relative to $L(\mathbf{Y})$ with $k$ satisfying the condition in the theorem, and we show that at least one of them satisfies conditions 2) and 3). Otherwise, if 2) is always violated, then there is an open path between some leaf node in $L(\mathbf{Y})$, denoted by $L(Y_k)$, and some component of $\mathbf{Z}$, denoted by $Z_j$, and this open path does not go through any common cause of the variables in $L(\mathbf{Y})$. Then they have some common cause that does not cause any other variable in  $L(\mathbf{Y})$. Consequently, there exists at least one noise term, denoted by $\varepsilon_i$, that contributes to both $L(Y_k)$ (and hence $Y_k$) and $Z_j$, but not any other variables in $\mathbf{Y}$. Because of the non-Gaussianity of the noise terms and Darmois-Skitovitch Theorem, if any linear projection of $\mathbf{Y}$, $\omega^\intercal \mathbf{Y}$ is independent from $\mathbf{Z}$, the linear coefficient for $Y_k$ must be zero. Hence $(\mathbf{Z}, \mathbf{Y}\setminus\{Y_k\})$ satisfies GIN, which contradicts the assumption in the theorem. Therefore, there must exists some $\mathcal{S}_{L}^k$ such that 2) holds. Next, if 3) is violated, i.e., the rank of the covariance matrix of $\mathcal{S}_{L}^k$ and $\mathbf{Z}$ is smaller than $k$. Then the condition 
$\omega^\intercal \mathbb{E}[\mathbf{YZ}^\intercal] = 0$ does not guarantee that $\omega^\intercal A = 0$. Under the faithfulness assumptions, we then do not have that $\omega^\intercal \mathbf{Y}$ is independent from $\mathbf{Z}$. Hence, condition 3) also holds. 
\end{proof}

\paragraph{Remark.} Roughly speaking, the conditions in this theorem can be interpreted the following way: i.) a causally earlier subset (according to the causal order) of the common causes of $\mathbf{Y}$ d-separate $\mathbf{Y}$ from $\mathbf{Z}$, and ii.) the linear transformation from that subset of the common causes to $\mathbf{Z}$ has full column rank. For instance, for the structure in Figure 1 of the main paper, $(\{X_3,X_4\}, \{X_1,X_2,X_5\})$ satisfies GIN, while $(\{X_3,X_6\}, \{X_1,X_2,X_5\})$ does not--note that the difference is that in the latter case one of the variables in $\mathbf{Z}$, $X_6$, is not d-separated from a component of $\mathbf{X}$, which is $X_5$, given the common causes of $\mathbf{X}$. However, when $X_6$ is replaced by $X_4$ in $\mathbf{Z}$, whose direct cause is causally earlier, the d-separation relationship holds, and so is the GIN condition. 

\subsection{Proof of Theorem 3}
\begin{reptheorem}{The-basic-cluster}
	Let $\mathbf{X}$ be the set of all observed variables in a LiNGLaM and $\mathbf{Y}$ be a proper subset of $\mathbf{X}$. If $(\mathbf{X} \setminus \mathbf{Y},\mathbf{Y})$ follows the GIN condition and there is no subset $\tilde{\mathbf{Y}} \subseteq \mathbf{Y}$ such that $(\mathbf{X} \setminus \tilde{\mathbf{Y}},\tilde{\mathbf{Y}})$ follows the GIN condition, then $\mathbf{Y}$ is a causal cluster and $Dim(L(\mathbf{Y}))=Dim(\mathbf{Y})-1$.
\end{reptheorem}
\begin{proof}
We will prove it by contradiction. Let $\mathbf{Y}=(Y_1^\intercal,...,Y_{Dim(\mathbf{Y})}^\intercal)^\intercal$. There are two cases to consider. 

Case 1). Assume that $\mathbf{Y}$ is not a causal cluster and show that $(\mathbf{X} \setminus {\mathbf{Y}},\mathbf{Y})$ violates the GIN condition, leading to the contradiction.
Since $\mathbf{Y}$ is not a causal cluster, without loss of generality, $L(\mathbf{Y})$ must contain at least two different parental latent variable sets, denoted by $L_a$ and $L_b$.
Now, we show that there is no non-zero vector $\omega$ such that ${\omega}^\intercal \mathbf{Y}$ is independent from $\mathbf{X} \setminus \tilde{\mathbf{Y}}$.
Because there is no subset $\tilde{\mathbf{Y}} \subseteq \mathbf{Y}$ such that $(\mathbf{X} \setminus \tilde{\mathbf{Y}},\mathbf{Y})$ follows the GIN condition, the number of elements containing the components of $L_a$ in $\mathbf{Y}$ is smaller than $Dim(L_a)+1$ and the number of elements containing the components of $L_b$ in $\mathbf{Y}$ is less than $Dim(L_b)+1$.
Thus, we obtain that there is no ${\omega}\neq 0$ such that ${\omega}^\intercal \mathbb{E}[\mathbf{Y}((\mathbf{X} \setminus \mathbf{Y})^\intercal]=0$. That is to say, ${\omega}^\intercal \mathbf{Y}$ is dependent on $\mathbf{X} \setminus \mathbf{Y}$, i.e., $(\mathbf{X} \setminus {\mathbf{Y}},\mathbf{Y})$ violates the GIN condition, which leads to the contradiction.

Case 2). Assume that $\mathbf{Y}$ is a causal cluster but $Dim(L(\mathbf{Y})) \neq Dim(\mathbf{Y})-1$. 
First, we consider the case where $Dim(L(\mathbf{Y})) < Dim(\mathbf{Y})-1$.
If $Dim(L(\mathbf{Y})) > Dim(\mathbf{Y})-1$, we always can find a subset $\tilde{\mathbf{Y}} \subseteq \mathbf{Y}$ and $Dim(\tilde{\mathbf{Y}})=Dim((L(\mathbf{Y}))+1$ such that $(\mathbf{X} \setminus \tilde{\mathbf{Y}},\mathbf{Y})$ follows the GIN condition, leading to the contradiction.

We then consider the case where $Dim(L(\mathbf{Y})) > Dim(\mathbf{Y})-1$.
Due to the linear assumption, each element in $L(\mathbf{Y})$ contains components $\{\varepsilon_{L^{Y}_{1}},...,\varepsilon_{L^{Y}_{Dim(L(\mathbf{Y}))}}\}$.
Because $Dim(L(\mathbf{Y})) > Dim(\mathbf{Y})-1$, ${\omega}^\intercal (Y_{1}^\intercal,....,Y_{Dim(\mathbf{Y})}^\intercal)^\intercal$ contains $\varepsilon_{L^{Y}_{i}}, i \in \{1,..,Dim(L(\mathbf{Y}))\}$, for any ${\omega} \neq 0$. According to the Darmois-Skitovitch Theorem, we have ${\omega}^\intercal (Y_{1}^\intercal,....,Y_{Dim(\mathbf{Y})}^\intercal)^\intercal \nupvDash \mathbf{X} \setminus {\mathbf{Y}}$. That is to say, $(\mathbf{X} \setminus {\mathbf{Y}},\mathbf{Y})$ violates the GIN condition, which leads to a contradiction.
\end{proof}

\subsection{Proof of Theorem 4}
\begin{reptheorem}{theo: the-basic-causa-direction}
Let $\mathcal{S}_p$ and $\mathcal{S}_q$ be two causal clusters of a LiNGLaM. Assume there is no latent confounder for $L(\mathcal{S}_p)$ and $L(\mathcal{S}_q)$, and $L(\mathcal{S}_p) \cap L(\mathcal{S}_q)= \emptyset$. Further suppose that $\mathcal{S}_p$ contains $2Dim(L(\mathcal{S}_p))$ number of variables with $\mathcal{S}_p=\{P_{1},P_{2},...,P_{2Dim(L(\mathcal{S}_p))}\}$ and that $\mathbf{C}_q$ contains $2Dim(L(\mathcal{S}_q))$ number of variables with $\mathcal{S}_q=\{Q_{1},Q_{2},...,Q_{2Dim(L(\mathcal{S}_q))}\}$. Then if $(\{P_{Dim(L(\mathcal{S}_p))
+1},...P_{2Dim(L(\mathcal{S}_p))}\},\{P_{1},....,P_{Dim(L(\mathcal{S}_p))},Q_{1},...Q_{Dim(L(\mathcal{S}_q))}\})$ follows the GIN condition, $L(\mathcal{S}_p) \to L(\mathcal{S}_q)$ holds. 
\end{reptheorem}
\begin{proof}
For $L(\mathcal{S}_p)$ and $L(\mathcal{S}_q)$, there are two possible causal relations: $L(\mathcal{S}_p) \to L(\mathcal{S}_q)$ and $L(\mathcal{S}_p) \leftarrow L(\mathcal{S}_q)$. 
For clarity, let $m=Dim(L(\mathcal{S}_p))$ and $n=Dim(L(\mathcal{S}_p))$. Further, Let $L(\mathcal{S}_p)=\{L^{p}_{1},...,L^{p}_{m}\}$ and $L(\mathcal{S}_q)=\{L^{q}_{1},...,L^{q}_{n}\}$ (note that subscripts denote the causal order).

First, we consider case 1: $L(\mathcal{S}_p) \to L(\mathcal{S}_q)$, by leveraging the result of Theorem  \ref{Theo-basic}. 

According to the linearity assumption, we have
\begin{align}\label{eq-theo4-Lp-Lq-Y}
\underbrace{\left[ {\begin{matrix}
{P_1}\\
 \vdots \\
{P_{m}} \\
{Q_1}\\
 \vdots \\
{Q_{n}} 
\end{matrix}} \right]}_{\mathbf{Y}} & = \underbrace{\left[ {\begin{matrix}
{{C_{11}}}&\cdots &{{C_{m1}}}\\
 \vdots & \ddots  &\vdots\\
{{C_{m1}}}& \cdots &{{C_{mm}}}\\
{{D_{11}}}&\cdots &{{D_{n1}}}\\
 \vdots & \ddots  &\vdots\\
{{D_{n1}}}& \cdots &{{D_{nn}}}
\end{matrix}} \right]}_{A}\underbrace{\left[ {\begin{matrix}
{L_1^p}\\
 \vdots \\
{L_m^p}
\end{matrix}} \right]}_{\mathbf{L}}+
\underbrace{\left[ {\begin{matrix}
{{\varepsilon _{P_1}}}\\
 \vdots \\
{{\varepsilon _{P_{m}}}}\\
{{\varepsilon' _{Q_1}}}\\
 \vdots \\
{{\varepsilon' _{Q_{n}}}}
\end{matrix}} \right]}_{\mathbf{E}_{Y}}
\end{align}
and
\begin{align}\label{eq-theo4-Lp-Lq-Z}
\underbrace{\left[ {\begin{matrix}
{P_{m+1}}\\
 \vdots \\
{P_{2m}} 
\end{matrix}} \right]}_{\mathbf{Z}} & = \underbrace{\left[ {\begin{matrix}
{{B_{11}}}&\cdots &{{B_{m1}}}\\
 \vdots & \ddots  &\vdots\\
{{B_{m1}}}& \cdots &{{B_{mm}}}\\
\end{matrix}} \right]}_{B}\underbrace{\left[ {\begin{matrix}
{L_1^p}\\
 \vdots \\
{L_m^p}
\end{matrix}} \right]}_{\mathbf{L}}+
\underbrace{\left[ {\begin{matrix}
{{\varepsilon _{P_{m+1}}}}\\
 \vdots \\
{{\varepsilon _{P_{2m}}}}
\end{matrix}} \right]}_{\mathbf{E}_{Z}},
\end{align}
where ${{\varepsilon' _{Q_i}}}=\sum\limits_{k = 1}^n f_k{\varepsilon_{L^{q}_{k}}}+{\varepsilon _{Q_i}}$.

Now, we verify conditions 1) $\sim$ 4) in Theorem  \ref{Theo-basic}.
Based on Equations \ref{eq-theo4-Lp-Lq-Y} and \ref{eq-theo4-Lp-Lq-Z}, we have $Dim(\mathbf{L})=m$.
For condition 1), $Dim(\mathbf{Y})=m+n>m$.
For condition 2), 
$\mathbf{E}_{Y}=({{\varepsilon _{P_1}}},...,{{\varepsilon _{P_{m}}}},{{\varepsilon' _{Q_1}}},...,{{\varepsilon' _{Q_n}}} )^\intercal$ is independent from $\mathbf{L}=\{L^{p}_{1},...,L^{p}_{m}\}$, due to the fact that there is no common component between $\mathbf{E}_{Y}$ and $\mathbf{L}$ and that each component is independent of each other. For condition 3), because $\varepsilon _{P_{k}},k=1,...,2m$, is independent from $\mathbf{L}$, $\mathbf{E}_{Z} \upvDash \mathbf{L}$.
For condition 4), $\boldsymbol{\Sigma}_{\mathbf{LZ}} = \mathbb{E}[\mathbf{L}\mathbf{Z}^\intercal]=\boldsymbol{\Sigma}_{L} B^\intercal$. Because $Dim(B)=m$, we obtain that $\boldsymbol{\Sigma}_{\mathbf{LZ}}$ has rank $m$.
Therefore, $(\{P_{m+1)
+1},...P_{2m}\},\{P_{1},....,P_{m},Q_{1},...Q_{n}\})$ follows the GIN condition.

Next, we consider case 2: $L(\mathcal{S}_p) \leftarrow L(\mathcal{S}_q)$. According to the definition of the GIN condition, 
we need to find a vector ${\omega}\neq 0$ such that ${\omega}^\intercal \mathbb{E}[{(P_{1},....,P_{m},Q_{1},...Q_{n})}(P_{m+1},....,P_{2m})^\intercal]=0$. 
Due to the linearity assumption, each element in $\{P_{1},....,P_{2m}\}$ contains the component in $\varepsilon_{L^{P}_{1}},...,\varepsilon_{L^{P}_{m}}$ while $\{Q_{1},...Q_{n}\}$ not.
Because the dimension of $\varepsilon_{L^{P}_{i}}$ in $\{P_{1},....,P_{m},Q_{1},...Q_{n}\}$ is $m$ and $Dim(L(\mathcal{S}_p))=m$, ${\omega}^\intercal (P_{1},....,P_{m},Q_{1},...Q_{n})$ contains $\varepsilon_{L^{P}_{i}}$, for any ${\omega} \neq 0$. According to the  Darmois-Skitovitch Theorem, we have ${\omega}^\intercal (P_{1},....,P_{m},Q_{1},...Q_{n}) \nupvDash (P_{m+1},....,P_{2m})^\intercal$. That is to say, $(\{P_{m+1)
+1},...P_{2m}\},\{P_{1},....,P_{m},Q_{1},...Q_{n}\})$ violates the GIN condition.

Therefore, $L(\mathcal{S}_p) \to L(\mathcal{S}_q)$.
\end{proof}

\subsection{Proof of Proposition 3}
\begin{repproposition}{property_merge}
    Let $\mathcal{S}_1$ and $\mathcal{S}_2$ be two clusters of a LiNGLaM and $Dim(L(\mathcal{S}_1))=Dim(L(\mathcal{S}_2))$. If $\mathcal{S}_1$ and $\mathcal{S}_2$ are overlapping, $\mathcal{S}_1$ and $\mathcal{S}_2$ share the same set of latent variables.
\end{repproposition}
\begin{proof}
Because $\mathcal{S}_1$ and $\mathcal{S}_2$ are overlapping, with loss of generality, assume that the shared element of $\mathcal{S}_1$ and $\mathcal{S}_2$ is $X_k$. Furthermore, we have that $L(\mathcal{S}_1)$ and $L(\mathcal{S}_2)$ are both parents of $X_k$. Based on the definition of causal cluster and that $Dim(L(\mathcal{S}_1))=Dim(L(\mathcal{S}_2))$, we have $L(\mathcal{S}_1)=L(\mathcal{S}_2)$. That is to say, $\mathcal{S}_1$ and $\mathcal{S}_2$ share the same set of latent variables.  
\end{proof}

\subsection{Proof of Lemma 1}
\begin{replemma}{lemma: Find-Causal-Order-Root-normal}
    Let $\mathcal{S}_r$ be a cluster and $\mathcal{S}_k,k \neq r$ be any other cluster of a LiNGLaM. Suppose that $\mathcal{S}_r$ contains $2Dim(L(\mathcal{S}_r))$ number of variables with $\mathcal{S}_r=\{R_{1},R_{2},...,R_{2Dim(L(\mathcal{S}_r))}\}$ and that $\mathcal{S}_k$ contains $2Dim(L(\mathcal{S}_k))$ number of variables with $\mathcal{S}_k=\{K_{1},K_{2},...,K_{2Dim(L(\mathcal{S}_k))}\}$.
    if $(\{R_{Dim(L(\mathcal{S}_r))+1},...R_{2Dim(L(\mathcal{S}_r))},\{R_{1},....,R_{Dim(L(\mathcal{S}_r))},K_{1},...K_{Dim(L(\mathcal{S}_k))}\})$ follows the GIN condition, then $L(\mathcal{S}_r)$ is a root latent variable set.
\end{replemma}
\begin{proof}
(i) Assume that $L(\mathcal{S}_r)$ is a root latent variable set. Due to the linearity assumption, there is no latent confounder between $L(\mathcal{S}_r)$ and another latent variable set. Based on Theorem \ref{theo: the-basic-causa-direction}, we have that $(\{R_{|L(\mathcal{S}_r)|},...R_{2Dim(L(\mathcal{S}_r))},\{R_{1},....,R_{Dim(L(\mathcal{S}_r))},K_{1},...K_{Dim(L(\mathcal{S}_k))}\})$ follows the GIN condition.

(ii) Assume that $L(\mathcal{S}_r)$ is not a root latent variable set, that is , $L(\mathcal{S}_r)$ has at least one parent set. Let $L(\mathcal{S}_p)$ be the parent of $L(\mathcal{S}_r)$ and $\mathcal{S}_p=\{P_{1},P_{2},...,P_{2Dim(L(\mathcal{S}_p))}\}$.
Thus, every element in $\{P_{1},P_{2},...,P_{2Dim(L(\mathcal{S}_p))}\}$ has the component $\varepsilon_{L(\mathcal{S}_p)}$.
Based on the definition of the GIN condition, we easily obtain that there is no ${\omega}\neq 0$ such that ${\omega}^\intercal \mathbb{E}[{\{P_{1},....,P_{Dim(L(\mathcal{S}_p))},R_{1},...R_{Dim(L(\mathcal{S}_r))}\})},({\{P_{Dim(L(\mathcal{S}_p))+1},...R_{2Dim(L(\mathcal{S}_p))}})^\intercal]=0$ because the dimension of $\varepsilon_{L(\mathcal{S}_p)}$ in $\{R_{1},....,R_{Dim(L(\mathcal{S}_r))},K_{1},...K_{Dim(L(\mathcal{S}_k))}\}$ equals $Dim(L(\mathcal{S}_r))$. That is to say, $\omega^\intercal (R_{1},....,R_{Dim(L(\mathcal{S}_r))},K_{1},...K_{Dim(L(\mathcal{S}_k))})$ must have the component $\varepsilon_{L(\mathcal{S}_p)}$. Thus, $\omega^\intercal (R_{1},....,R_{Dim(L(\mathcal{S}_r))},K_{1},...K_{Dim(L(\mathcal{S}_k))})$ is dependent on $(P_{Dim(L(\mathcal{S}_p))+1},...R_{2Dim(L(\mathcal{S}_p))})^\intercal$ based on the  Darmois-Skitovitch Theorem. Therefore, $(\{P_{Dim(L(\mathcal{S}_p))+1},...P_{2Dim(L(\mathcal{S}_p))}\},\{P_{1},....,P_{Dim(L(\mathcal{S}_p))},R_{1},...R_{Dim(L(\mathcal{S}_r))}\})$ violates the GIN condition.

From (ii), the lemma is proven. Moreover, from (i) and (ii), we show that   $(\{R_{|L(\mathcal{S}_r)|},...R_{2Dim(L(\mathcal{S}_r))}\},\{R_{1},....,R_{Dim(L(\mathcal{S}_r))},K_{1},...K_{Dim(L(\mathcal{S}_k))}\})$ follows the GIN condition, if and only if $L(\mathcal{S}_r)$ is a root latent variable set.
\end{proof}
\subsection{Proof of Proposition 4}
\begin{repproposition}{proposition: Find-Causal-Order-Root-normal}
    Suppose that $\{\mathcal{S}_{1},...\mathcal{S}_{i},...,\mathcal{S}_{n}\}$ are all the clusters of the LiNGLaM. Denote $\mathbf{T}=\{L(\mathcal{S}_{1}),...L(\mathcal{S}_{i})\}$ and $\mathbf{T}=\{L(\mathcal{S}_{i+1}),...L(\mathcal{S}_{n})\}$, where all elements in $\mathbf{T}$ are causally earlier than those in $\mathbf{R}$. Let $\mathbf{{\hat Z}}$ contain the elements from the half children of each latent variable set in $\mathbf{T}$, and $\mathbf{{\hat Y}}$ contain the elements from the other half children of each latent variable set in $\mathbf{T}$. Furthermore, 
    Let $L(\mathcal{S}_r)$ be a latent variable set of  $\mathbf{R}$ and $\mathcal{S}_r=\{R_{1},R_{2},...,R_{2Dim(L(\mathcal{S}_r))}\}$. If for any one of the remaining $\mathcal{S}_k \in \mathbf{R},k \neq r$ and $\mathcal{S}_k=\{K_{1},K_{2},...,K_{2Dim(L(\mathcal{S}_k))}\}$ such that $(\{R_{Dim(L(\mathcal{S}_r))+1},...,R_{2Dim(L(\mathcal{S}_r))},\mathbf{{\hat Z}}\},\{R_{1},....,R_{Dim(L(\mathcal{S}_r))},K_{1},...K_{Dim(L(\mathcal{S}_k))},\mathbf{{\hat Y}}\})$ follows GIN condition, then $L(\mathcal{S}_r)$ is a root latent variable set in $\mathbf{R}$.
\end{repproposition}
\begin{proof}
One may treat the causally earlier sets as a new group. Then one can easily prove this result according to Lemma \ref{lemma: Find-Causal-Order-Root-normal}.
\end{proof}

\section{More experimental results of Synthetic data}
Here, we add more results to show the performance of our algorithm for random generated graphs and more variables. In details, we generated graphs randomly with different numbers of latent variables, where each latent variable only have three observed variables. We run our method and obtain the following results in Table \ref{Fig:more_experiments}.
\begin{table*}[htp]\scriptsize
	\small
	\center \caption{results with different numbers of variables and randomly generated graphs (with sample size=2000).}
	\vspace{-2ex}
	\label{Fig:more_experiments}
	\resizebox{\textwidth}{8mm}{ 
	\begin{tabular}{|c|c|c|c|c|}
		\hline  \multicolumn{1}{|c|}{\textbf{Number of variables (latent variables)}}
		&\multicolumn{1}{|c|}{\textbf{Latent omission}} & \multicolumn{1}{|c|}{\textbf{Latent commission}} & \multicolumn{1}{|c|}{\textbf{Mismeasurements}}&
		\multicolumn{1}{|c|}{\textbf{Correct-ordering rate}}\\ \hline 
		\multicolumn{1}{|c|}{15(5)}  & 0.02(1) & 0.00(0) & 0.00(0) & 0.90 \\\hline 
		\multicolumn{1}{|c|}{30(10)} & 0.09(3) & 0.05(3) & 0.04(3) & 0.85  \\\hline 
		\multicolumn{1}{|c|}{60(20)}  & 0.15(6) & 0.12(6) & 0.10(6) & 0.79  \\\hline 
	\end{tabular}}
\end{table*}
\section{More details of Real-Word data}
For comparisons, we give the hypothesized factors formulated in \citep{byrne2010structural} in Table \ref{tab:causal-clusters-s}.
\begin{table}[htp!]
\setlength\tabcolsep{0pt}
\small
    \centering
    \begin{tabular}{|c|c|}
\hline
Factors & Observed variables \\ \hline
\emph{Role Conflict} & $RC_{1}$, $RC_{2}$, $WO_{1}$, $WO_{2}$,  \\\hline
\emph{Decision Making} & $DM_{1}$, $DM_{2}$\\\hline
\emph{Classroom Climate} & $CC_{1}$, $CC_{2}$,$CC_{3}$,$CC_{4}$\\\hline
\emph{Self-Esteem} & $SE_{1}$, $SE_{2}$,$SE_{3}$\\\hline
\emph{Peer Support} & $PS_{1}$, $PS_{2}$\\\hline
~\emph{External Locus of Control}~ & ~~$ELC_{1}$, $ELC_{2}$,$ELC_{3}$,$ELC_{4}$,$ELC_{5}$~~\\\hline
\emph{Emotional Exhaustion} & $EE_{1}$, $EE_{2}$,$EE_{3}$\\\hline
\emph{Denationalization}  & $DP_{1}$, $DP_{2}$, $DP_{3}$ \\ \hline
\emph{Personal Accomplishment}  & $PA_{1}$, $PA_{2}$, $PA_{3}$ \\
\hline
    \end{tabular}
    \caption{The hypothesized factors in \citep{byrne2010structural}.}
    \label{tab:causal-clusters-s}
\end{table}

\medskip

\small

\bibliographystyle{plainnat}
\bibliography{reference}

\end{document}